\documentclass{applemlr}

\usepackage{amsmath}
\usepackage{enumerate}
\usepackage{algorithm}
\usepackage{algpseudocode}
\usepackage{amsfonts}
\usepackage{amsthm}
\usepackage{cleveref}
\usepackage{diagbox}
\usepackage{colortbl}
\usepackage{amssymb}
\usepackage{xspace}
\usepackage{wrapfig}
\usepackage{adjustbox}
\usepackage{tabularx}
\usepackage{booktabs}
\usepackage{mathtools}
\usepackage{tikz}
\usepackage{enumitem}
\usepackage{silence}
\usepackage{dsfont}
\usepackage[table]{xcolor}
\usepackage[dvipsnames]{xcolor}
\usepackage{multirow}
\usepackage{makecell}
\usepackage{xfakebold}

\usepackage{amsmath,amsfonts,bm}

\def\eqref#1{equation~\ref{#1}}

\def\1{\bm{1}}

\DeclareMathAlphabet{\mathsfit}{\encodingdefault}{\sfdefault}{m}{sl}
\SetMathAlphabet{\mathsfit}{bold}{\encodingdefault}{\sfdefault}{bx}{n}

\DeclareRobustCommand\onedot{\futurelet\@let@token\@onedot}
\def\onedot{.} %

\def\etal{\emph{et al}\onedot~}

\newcommand{\mytilde}{\raise.17ex\hbox{$\scriptstyle\mathtt{\sim}$}}

\newcommand{\setL}{\mathcal{L}}
\newcommand{\setU}{\mathcal{U}}
\newcommand{\setS}{\mathcal{S}}

\newcommand{\serverD}{\setL}
\newcommand{\clientD}{\setU}
\newcommand{\teacher}{\phi}
\newcommand{\student}{\theta}
\newcommand{\local}{\bar{\theta}}
\newcommand{\clients}{\setS}
\definecolor{textgray}{HTML}{6E6E73}
\usetikzlibrary{positioning, calc}
\usetikzlibrary{decorations.pathmorphing}

\makeatletter
\patchcmd{\wrong@fontshape}{\@gobbletwo}{}{}{}
\makeatother
\numberwithin{equation}{section}
\makeatletter
\AtBeginDocument{
  \urlstyle{sf}
  
}
\makeatother

\definecolor{light}{RGB}{125, 125, 125}
\crefname{tcb@cnt@pbox}{code}{code}
\Crefname{tcb@cnt@pbox}{Code}{Code}
\crefname{assumption}{assumption}{assumption}
\Crefname{assumption}{Assumption}{Assumptions}

\newtcolorbox[auto counter]{pbox}[2][]{
  colback=white,
  title=Code~\thetcbcounter: #2,
  #1,fonttitle=\sffamily,
  fontupper=\sffamily,
  arc=2pt,
  colframe=bgcolor,
  coltitle=fgcolor,
  colbacktitle=bgcolor,
  toptitle=0.25cm,
  bottomtitle=0.125cm
}

\makeatletter
\newcommand\applefootnote[1]{%
  \begingroup
  \renewcommand\thefootnote{}%
  \renewcommand\@makefntext[1]{\noindent##1}%
  \footnote{#1}%
  \addtocounter{footnote}{-1}%
  \endgroup
}
\makeatother

\definecolor{cverbbg}{gray}{0.90}

\newif\ifapplebuild
\applebuildfalse

\setcitestyle{authoryear,round,citesep={;},aysep={,},yysep={;}}

\usepackage{xcolor}
\usepackage{hyperref}
\usepackage{cleveref}
\usepackage{subcaption}
\usepackage{url}

\usepackage{algorithm}
\usepackage{algpseudocode}
\algrenewcommand\algorithmicrequire{\textbf{Input:}}
\algrenewcommand\algorithmicensure{\textbf{Output:}}
\usepackage{booktabs}
\usepackage{multirow}
\usepackage{hhline}
\usepackage{makecell}
\usepackage{bigints}
\usepackage{nicematrix}
\usepackage{tabularx}
\usepackage{fontawesome5}
\usepackage[most]{tcolorbox}
\usepackage{enumitem}
\usepackage{wrapfig}

\newtcolorbox{takeaways}{
  colback=black!5,
  colframe=black!50,
  boxrule=0.5pt,
  arc=2pt,
  left=8pt, right=8pt, top=6pt, bottom=6pt,
  title=\textbf{Take-aways},
  coltitle=black,
  colbacktitle=black!15,
  fonttitle=\bfseries,
}

\definecolor{mpl_green}{HTML}{2ca02c}
\definecolor{mpl_blue}{HTML}{1f77b4}
\definecolor{mpl_orange}{HTML}{ff7f0e}
\definecolor{mpl_red}{HTML}{d62728}

\definecolor{pastelred}{RGB}{255, 105, 97}

\definecolor{vOneEditColor}{named}{black}
\newcommand{\vOneEdits}[1]{{\color{vOneEditColor}#1}}

\definecolor{vTwoEditColor}{named}{black}
\newcommand{\vTwoEdits}[1]{{\color{vTwoEditColor}#1}}

\definecolor{vThreeEditColor}{named}{black}
\newcommand{\vThreeEdits}[1]{{\color{vThreeEditColor}#1}}

\usepackage[colorinlistoftodos,textsize=tiny,disable]{todonotes}
\applebuildtrue

\counterwithout{equation}{section}

\title{A Practical Recipe for Semi-Supervised Federated ASR:
Online Pseudo-Labels with Server Update Stabilization}

\author{Wonho~Bae}
\author{Zakaria~Aldeneh}
\author{Martin~Pelikan}
\author{Jan~``Honza''~Silovsky}
\author{Tatiana~Likhomanenko}
\author{Sheikh~Shams~Azam}

\affiliation{Apple}

\abstract{
Semi-supervised federated learning (SSFL) trains models on clients' unlabeled data using a teacher to generate pseudo-labels, with a small labeled seed dataset on the server. Automatic Speech Recognition (ASR) is particularly fragile here: pseudo-label errors compound across the output sequence and across training rounds into divergence, leaving a large gap to fully-supervised FL. 
We show that closing this gap turns on two coupled design axes---\textit{the teacher} (which model generates the pseudo-labels) and \textit{the anchor} (the server-side updates on labeled data that stabilize training). 
On the teacher axis, a per-client online teacher (each client's own evolving model) diverges on its own, but once stabilized it matches or beats the broadcast global teacher (one server model, fixed within a round)---decisively in-domain and competitively under domain shift.
As the seed grows stronger and the online teacher's advantage narrows, a transitioning teacher (global $\rightarrow$ online at round $r$) matches or beats both.
On the anchor axis, the server must keep training on labeled data between rounds---otherwise the online teacher drifts---and this interleaving, more than the seed model, governs convergence.
The two axes are inseparable: aggressive teacher choices pay off only once the anchor stabilizes training, which is highly sensitive to data augmentation and batch size---the settings that govern how much input and gradient noise the server injects.
How much stabilization is needed is domain-dependent, governed by the dispersion of the seed data and its overlap with client data.
These findings yield guidelines for SSFL in ASR training, improving over the strongest prior method on 9 of 11 pairs, by $20.8\%$ on average in-domain and $10.0\%$ cross-domain, narrowing the gap to fully-supervised FL.
}

\metadata[Correspondence]{\sffamily Sheikh Shams Azam: \href{mailto:s_azam@apple.com}{s\_azam@apple.com}}
\date{\sffamily\today}

\begin{document}

\maketitle

\applefootnote{\vOneEdits{We study semi-supervised federated ASR entirely in simulation on public datasets, spanning a wide range of speaking styles and domain-shift conditions along with a broad sweep of training configurations, to distill practical guidelines. No real user data or production telemetry informed this work. All reported system parameters are choices made for these simulations and do not represent any currently deployed system or plan thereof.}}

%% >>>>> begin body.tex
\section{Introduction}
\label{sec:introduction}

Safeguarding privacy is a foundational commitment when building modern machine learning systems, especially when training on data that may be considered personal~\citep{truong2020fl_gdpr}. One option for alleviating such concerns is using synthetic data, which can be generated at scale with little to no privacy concerns. However, synthetic alternatives still face challenges and do not yet close the gap to real user data. For example, in speech, Text-to-Speech (TTS) substitutes struggle to replicate personalized speech patterns, diverse acoustic environments, and natural device characteristics, leaving distributional mismatch with real-world conditions~\citep{hilmes2024effect,su2024task,ogun2025exhaustive}. An alternative remedy---and the focus of this work---is federated learning \citep{truong2020fl_gdpr,jeong2024fed_unlearning_survey}.

Federated learning (FL)~\citep{mcmahan2017fl} inverts the pipeline---instead of moving data to the model, the model moves to the data, with user data never leaving the device.
Each device computes a model update locally on its own data and communicates only that update to the server, which aggregates the updates from many devices into a new global model.
Speech is a natural fit for this framework: user audio is abundant on consumer devices, privacy-sensitive when centralized, and not yet replaceable by synthetic alternatives.
Recent work has demonstrated that FL for end-to-end Automatic Speech Recognition (ASR) is feasible, with FL models approaching centrally-trained baselines~\citep{pelikan2023dp_fl_asr,guliani2021fl_asr}.

To strengthen privacy further, differential privacy (DP) adds calibrated noise to the communicated updates so that information unique to any single user is masked, bounding how much the shared updates can reveal about that user's data~\citep{abadi2016dpsgd}.
DP has been combined with FL for ASR via per-layer clipping techniques that mitigate DP noise's disproportionate impact on attention layers, achieving strong privacy guarantees at modest \vTwoEdits{word error rate (WER)} cost~\citep{pelikan2023dp_fl_asr}.
While this private deployment scenario motivates our work, we do not add DP noise in this first study: our recipe depends on large, low-variance server-side updates on labeled data to stabilize training against pseudo-label noise, and DP-SGD's per-layer clipping and noise injection may work against that variance reduction in ways that require dedicated tuning and analysis (\Cref{sec:discussion}); we therefore leave a joint study of DP and pseudo-labeling to future work.

Beyond privacy, a separate obstacle limits FL for ASR in practice: most work assumes that clients hold labeled data \citep{li2020fedprox, reddi2021fedopt, azam2022towards}---an assumption that rarely holds, as transcription is expensive and time consuming on device.
\emph{Semi-supervised federated learning} (SSFL) addresses this gap: the server holds a small labeled corpus---the \emph{seed data}---while clients contribute only unlabeled audio---the \emph{client data}. The two may also differ in distribution, with the seed often drawn from a curated corpus and client data reflecting diverse real-world speaking styles and acoustic conditions, making domain shift both between seed and clients, and across clients themselves a central concern.

This domain mismatch is further amplified by the nature of semi-supervised learning: \vTwoEdits{automated transcripts produced for a client's unlabeled audio---pseudo-labels---}inject additional noise into training, compounding the instability that domain shift already creates.
Therefore, for SSFL for ASR, two design choices are central: \textbf{which model generates the pseudo-labels} and \textbf{how the server ``anchors'' training using the labeled seed data}.\footnote{We use standard teacher--student terminology: the model being optimized is the \emph{student} \vTwoEdits{(each client's local model)} and the model that generates its pseudo-labels is the \emph{teacher}. The two may or may not be the same model.}
For pseudo-label generation, existing SSFL for ASR works use either a \emph{static teacher} (frozen at the seed model)~\citep{mehmood2022fednst} or a \emph{global teacher} (the broadcast model held fixed within each round but refreshed across rounds)~\citep{rao2023fl_self}, and combines server-side gradients with the aggregated client pseudo-gradients to reduce drift away from the labeled distribution.
The alternative, an \emph{online teacher} (each client's evolving local model, which is the student itself), has not been examined in the SSFL for ASR literature, as \citet{diao2022semifl} argue that the online teacher would drift from the global model and destabilize training.
Whether these defaults are optimal, or even adequate for ASR tasks, has not been systematically studied.

Across both axes, we observe that the conventional defaults turn out to be suboptimal, leaving a substantial WER gap to fully-supervised FL.
On the pseudo-label-source axis (described in \Cref{sec:pl-generation}), the online teacher matches or beats the global teacher in-domain and on most cross-domain settings.
As the strength of the seed model grows, the online--global gap shrinks, and a \emph{transitioning} teacher (global $\rightarrow$ online at round $r$) matches or beats both pure strategies when the global is better than the online teacher.
However, the online and transitioning teachers can break asymmetrically: swapping which corpus serves as the seed and which as the client can flip training from diverging to converging---evidence that the domain gap acts asymmetrically rather than as a symmetric distance between corpora.

In \Cref{sec:server-update}, we trace this failure to two server-update factors---the SpecAugment~\citep{park2019specaugment} (a data augmentation that masks blocks of time and frequency in the input features) strength and batch size used during server updates on the seed data.
Strong SpecAug causes outright divergence, with a deletion-only error signature in the middle of training.
Small batch does not necessarily diverge but destabilizes training by injecting noise into the server update, converging to a significantly higher WER than large batch.
Tuning both factors stabilizes SSFL training and substantially improves WER for both teachers, most dramatically for the online teacher, which it rescues from divergence.
In \Cref{sec:analysis}, we provide a holistic comparison against three existing SSFL for ASR methods across several source $\times$ target pairs, showing that the resulting recipe closes the gap to fully-supervised FL while providing practical guidance for both design choices.
Our contributions are summarized as follows:

\begin{enumerate}
    \item We analyze the choice of teacher for pseudo-labeling in SSFL for ASR---global, online, and transitioning. Contrary to the prevailing assumption that the online (per-client evolving) teacher is unstable, we show that it matches or beats the global teacher on most data pairs, and a transitioning teacher (global $\rightarrow$ online at round $r$) extends this advantage as the seed model grows stronger\vThreeEdits{~(\Cref{sec:pl-generation})}.
    \item We identify frequent enough server updates on the seed data as a key stability condition for the online and transitioning teachers\vThreeEdits{~(\Cref{sec:server-update})}.
    \item We characterize an asymmetric breakage of the online and transitioning teachers under cross-domain shift: swapping which corpus serves as the seed and which as the client can flip training from diverging to converging---evidence that the domain gap acts asymmetrically rather than as a symmetric distance between corpora\vThreeEdits{~(\Cref{subsec:online-breaks})}.
    \item \vTwoEdits{We analyze the two server-update factors that govern stability---SpecAug strength and batch size. Together they set the input and stochastic gradient noise each server update injects, and thus the cross-client consistency of the pseudo-labels; strong SpecAug causes training to diverge and a small batch destabilizes it, whereas tuning both stabilizes SSFL training}\vThreeEdits{~(\Cref{subsec:why-breaks-server})}.
    \item We compare the resulting recipe against three existing SSFL for ASR methods (Static PL, FedNST~\citep{mehmood2022fednst}, \citet{rao2023fl_self}) and a strong global teacher baseline derived from \citet{diao2022semifl} across several \vTwoEdits{source (server data) $\rightarrow$ target (client data) pairs}, showing that it closes the gap to fully-supervised FL\vThreeEdits{~(\Cref{subsec:comprehensive-comparison})}.
    \item We provide extensive ablation studies \vTwoEdits{characterizing how} cohort size, local steps, and the \vTwoEdits{choice of pseudo-label generator, i.e., teacher (\Cref{tab:teachers}) affect WER and training stability}\vThreeEdits{~(\Cref{subsec:ablations})}.
\end{enumerate}

\section{Related Work}
\label{sec:related-work}

\subsection{Federated Learning}

Federated learning (FL) enables collaborative model training across distributed clients without sharing raw data~\citep{mcmahan2017fl}, with \textsc{FedAvg} serving as the foundational algorithm that performs multiple local SGD steps before server-side aggregation.
A central challenge is \emph{client drift}: when client data is non-IID, local models diverge from the global optimum, slowing convergence~\citep{kairouz2021fl_survey}.
A broad line of work addresses this through proximal regularization~\citep{li2020fedprox}, control variates~\citep{karimireddy2020scaffold}, dynamic regularization~\citep{acar2021feddyn}, adaptive server-side optimization~\citep{reddi2021fedopt,wang2020fednova}, and personalization~\citep{fallah2020perfedavg}, all in the supervised setting.
Our work targets the semi-supervised FL setting, where the dominant source of instability is not client drift but \emph{pseudo-label drift}---the divergence of client-generated pseudo labels from the global model's pseudo labels---which we study along two axes: how pseudo labels are generated (\Cref{subsec:pl}) and how the server stabilizes training (\Cref{subsec:anchoring}).

\subsection{Pseudo-labeling: from Centralized SSL to SSFL}
\label{subsec:pl}

Pseudo-labeling---training on model-generated labels for unlabeled data---is the dominant approach to semi-supervised learning~\citep{lee2013pseudolabel}.
In image classification, FixMatch combines confidence-thresholded pseudo-labels with consistency regularization~\citep{sohn2020fixmatch}. Follow-up work refines the thresholding strategy via curriculum learning~\citep{zhang2021flexmatch} and self-adaptive schedules~\citep{wang2023freematch}, and recent analysis explains FixMatch's generalization advantage over supervised training~\citep{li2024safixmatch}.

For speech recognition, noisy student training generates pseudo-labels from a teacher model and iteratively retrains a larger student~\citep{park2020nst}; SlimIPL~\citep{likhomanenko2021slimipl} and momentum pseudo-labeling~\citep{higuchi2021mpl} replace the costly iterative regime with on-the-fly pseudo-label regeneration during training, and continuous pseudo-labeling extends this across the full training trajectory~\citep{likhomanenko2022cpl}.
A common thread is that the \emph{choice of pseudo-label source}---a fixed seed teacher, an iterative EMA teacher (which we call the \textit{global teacher}), or the evolving student itself (the \textit{online teacher})---is a central design axis.

In federated SSL, pseudo-labels are typically generated by the global server model broadcast each round, as in SemiFL~\citep{diao2022semifl} and most subsequent image-classification SSFL work.
FedSwitch~\citep{zhao2024fedswitch} is the closest prior work to study the choice of pseudo-label source: it adaptively switches between a global teacher and a teacher-student EMA based on relative quality, reporting that the choice substantially affects convergence.
BSemiFL~\citep{wang2025bsemifl} weighs the two sources via a Bayesian density estimator.
However, both approaches target image classification and assume discrete label spaces; FedSwitch additionally assumes clients are selected repeatedly across rounds so that per-client teacher state persists.
Neither assumption holds for sequence-output ASR with large, non-persistent client populations.

Other SSFL directions address orthogonal design choices: parameter decomposition for disjoint labeled/unlabeled aggregation~\citep{zhang2021fedmix}, consistency regularization~\citep{yang2023fedil,malaviya2023fedfame}, and knowledge-enhanced or prototype-based aggregation~\citep{wang2023knowledge,hu2026fed_arpl}.
These directions are complementary to the pseudo-labeling and anchoring questions studied here.

\subsection{Server Training and Anchoring}
\label{subsec:anchoring}

A parallel concern in semi-supervised learning is how to prevent the pseudo-label source from drifting as the model updates on its own predictions.
Mean Teacher maintains an exponential moving average of the student's weights as the teacher, anchoring pseudo-label generation to a slower-moving checkpoint~\citep{tarvainen2017meanteacher}.
Temporal ensembling achieves a similar effect by ensembling past predictions~\citep{laine2017temporal}.
Noisy Student iteratively re-initializes a larger student from a frozen teacher, which acts as a coarse-grained anchor between rounds of self-training~\citep{xie2020noisystudent_image}.
In the federated setting, the analogous question is how the server anchors training against noisy client pseudo-labels.

Federated SSL approaches vary in how the server stabilizes training.
FedNST freezes pseudo-labels generated once by the seed model, so the anchor is the seed itself~\citep{mehmood2022fednst}.
SemiFL alternates client pseudo-labeled rounds with server-side fine-tuning on the labeled seed data, using the labeled pass as a periodic anchor~\citep{diao2022semifl}.
\citet{rao2023fl_self} augment each round with a supervised gradient computed on a small held-out set of labeled server clients.
GDST integrates server-side fine-tuning to stabilize the global model against noisy client updates with client-side training on unlabeled data via pseudo-labeling and global distillation for image classification~\citep{liu2021gdst}.
Each of these strategies implicitly makes a different choice about when and how the labeled signal re-anchors training, but the implications for ASR---where pseudo-label quality is harder to assess and token-level errors compound---have not been systematically studied.

\subsection{Federated Learning in ASR}
Supervised FL for ASR is well-established.
\citet{guliani2021fl_asr} demonstrates FedAvg-style training for end-to-end ASR at scale, and \citet{pelikan2023dp_fl_asr} establish the first benchmark for differentially private FL-ASR using per-layer clipping to handle transformer gradient heterogeneity.
All of these assume fully labeled clients, which is impractical for on-device transcription.
The semi-supervised setting we study---where labels reside only at the server---is formalized next (\Cref{sec:ssfl-asr}), alongside the pseudo-labeling and server-anchoring questions introduced above.

\section{Semi-Supervised FL for ASR: Problem Setting and Practical Constraints}
\label{sec:ssfl-asr}

Semi-supervised FL admits four canonical configurations depending on where labeled data resides~\citep{song2024ssfl_survey}: \emph{labels-at-all-clients} (every client is labeled), \emph{labels-at-partial-clients} (only a subset of clients are labeled), \emph{unlabeled-at-server} (clients are labeled, the server is not), and \emph{labels-at-server} (the server is labeled, clients are not).
Of these, \emph{labels-at-server} setting is the closest to the \vOneEdits{constraints that motivate} deployed ASR: transcription is expensive and infeasible on-device, whereas service providers can curate a small, high-quality labeled corpus centrally.
The other three presuppose client-side labeling at a scale that is not available in practice.
Accordingly, \textbf{labels-at-server is the focus of this paper}, and it is also the setting adopted by all existing SSFL for ASR work~\citep{mehmood2022fednst,rao2023fl_self}.

A second distinction from centralized semi-supervised ASR is that the objective here is \emph{twofold}: (i) reduce WER on the target client domain by exploiting the unlabeled client audio, and (ii) avoid regressing on the server-side seed distribution, which represents the curated baseline the service is already committed to.
A strategy that improves client-domain WER at the cost of significant seed-corpus regression is not a practical win, so we report WER on both the target and the seed test sets throughout.

Compared to image classification, ASR also poses challenges that are specific to the output space: predictions are variable-length sequences rather than fixed-label classes, pseudo-label quality is harder to assess, and token-level errors compound at the sequence level.
As a result, only a handful of works have addressed SSFL for ASR.
FedNST~\citep{mehmood2022fednst} extends noisy student training~\citep{park2020nst} to the federated setting, generating pseudo-labels once with a seed model and never updating them throughout training.
\citet{rao2023fl_self} employ the global model to generate pseudo-labels for local updates as in~\citep{diao2022semifl}, with server-side gradients computed on a small held-out labeled subset.
Both approaches rely on a fixed or global teacher and do not explore locally updated teachers, leaving open the question of whether adaptive local pseudo-labeling can improve performance---the central question of our work.
We next formalize the training procedure and identify the practical constraints that shape the design space.

\begin{algorithm*}[t]
\caption{Semi-Supervised Federated Learning}
\label{alg:fl_slimipl}
\begin{algorithmic}[1]

\Require Model parameter $\theta$, set of clients $\clients$ with per-round cohort $\clients_t$, labeled server data $\serverD$, unlabeled client data $\clientD_i$ of client $i$, communication rounds $T$, server training probability $p$, server optimizer \textsc{ServerOpt}
\Ensure Trained global model parameters

\State \textit{\textbf{Seed Training Stage}}
\State Train a model on $\serverD$ until convergence to obtain $\theta_0$, used to initialize teacher $\teacher$ and student models $\student$

\Statex
\State \textit{\textbf{Federated Training Stage}}
\For{each communication round $t = 1, 2, \dots, T$}
    \State Server selects a subset of clients $\clients_t$ from $\clients$
    
    \ForAll{client $i \in \mathcal{S}_t$ in parallel}
        \State {\color{mpl_red}\textbf{PL generation:}} Generate pseudo-labels for $\clientD_i$ using teacher $\teacher_t$ (global) or $\student_{t,\tau}$ (online)
        \State \textbf{Data Filtering:} Remove samples with high uncertainty before local training
        \State \textbf{Local Training:} Optimize locally to $\student_t^{i}$ and send pseudo-gradient $g_t^{i} = \student_t - \student_t^{i}$ to the server
    \EndFor

    \State \textbf{FL update:} Aggregate $\bar{g}_t = \frac{1}{|\clients_t|}\sum_{i \in \clients_t} g_t^{i}$ (FedAvg) and update $\student_{t+1} = \textsc{ServerOpt}(\student_t, \bar{g}_t)$
    \State {\color{mpl_blue}\textbf{Server update:}} With probability $p$, further update $\student_{t+1}$ with a supervised step on $\serverD$
    \State {\color{mpl_red}\textbf{Teacher update:}} Update the teacher model $\teacher_t$ if applicable, e.g., EMA
\EndFor

\end{algorithmic}
\end{algorithm*}

\begin{table}[t!]
\centering
\small
\begin{tabular}{l c c c c}
\toprule
 & \textbf{Static PL} & \textbf{FedNST} & \textbf{Rao \etal} & \textbf{Diao \etal for ASR} \\
\midrule
{\color{mpl_red}\textbf{PL generation}} & one-time $\theta_0$& one-time $\theta_0$ & EMA teacher $\teacher_t$ & EMA teacher $\teacher_t$ \\
{\color{mpl_blue}\textbf{Server updates}} & no server updates & full server data & held-out server clients & a batch of server data \\
\bottomrule
\end{tabular}
\caption{Comparison of SSFL methods along the pseudo-label generation and server training axes.}
\label{tab:method_comparison}
\end{table}

\begin{table}[t]
\centering
\small
\begin{tabular}{ll}
\toprule
\textbf{Teacher} & \textbf{Pseudo-label source (at round $t$, local step $\tau$)} \\
\midrule
Global        & \vOneEdits{EMA of the broadcast model, $\teacher_t = \lambda\teacher_{t-1}+(1-\lambda)\student_t$ ($\lambda{=}0.99$), held fixed within the round} \\
Online        & client's local model $\student_{t,\tau}$, evolving within the round \\
Local EMA     & $\local_{t,\tau} = \gamma\,\local_{t,\tau-1} + (1-\gamma)\,\student_{t,\tau}$ \;($\gamma{=}1$: global, $\gamma{=}0$: online) \\
Transitioning & global teacher for the first $r$ rounds, online teacher thereafter \\
\bottomrule
\end{tabular}
\caption{The pseudo-label teachers compared in this paper. The global and online teachers are the two primary sources; the local EMA teacher (\Cref{subsec:global-vs-online}) interpolates between them, and the transitioning teacher (\Cref{subsec:seed-strength}) switches from global to online at round $r$. \vOneEdits{Note that the global teacher's cross-round decay $\lambda$ and the local EMA teacher's within-round decay $\gamma$ (reset each round) are independent parameters on different timescales.}}
\label{tab:teachers}
\end{table}

\subsection{Problem Setting}
\label{subsec:problem-setting}

\paragraph{Training template.}
Algorithm~\ref{alg:fl_slimipl} summarizes the SSFL training procedure under the labels-at-server setting.
Training proceeds in two stages.
In the \emph{seed stage}, the server trains a model on the labeled corpus $\serverD = \{(\mathbf{x}_j, \mathbf{y}_j)\}_{j=1}^{N_s}$ until convergence, obtaining a model parameter $\theta_0$; this seed model initializes both the student (the model being trained) and the teacher (the model that generates pseudo-labels).
In the \emph{federated stage}, each communication round $t$ proceeds as follows: (i) server broadcasts the current global model to a cohort $\clients_t \subset \clients$, (ii) each client~$i$ generates pseudo-labels for its unlabeled audio $\clientD_i = \{\mathbf{x}_{i,j}\}_{j=1}^{N_C}$ using some teacher, optionally filters uncertain samples, applies augmentation, and performs several local SGD steps, (iii) clients send their resulting pseudo-gradients back to the server, which aggregates them (e.g., with FedAvg~\citep{mcmahan2017fl}) and applies a server update, and (iv) model is updated further with the labeled seed data before updating the teacher.

\paragraph{Design axes.}
Existing SSFL methods differ along two axes within this template: the \textbf{pseudo-label source} (\Cref{subsec:pl})---which model generates the pseudo-labels for client training---and the \textbf{server training strategy} (\Cref{subsec:anchoring})---how and when the server re-injects the labeled seed signal into the global model.
Table~\ref{tab:method_comparison} summarizes how existing methods instantiate these two axes.

\paragraph{Prior methods.}
Along the \emph{pseudo-label source} axis, Static PL and FedNST~\citep{mehmood2022fednst} generate pseudo-labels once from the initial seed model~$\theta_0$ and keep them fixed throughout training, whereas \citet{rao2023fl_self} and \citet{diao2022semifl} regenerate pseudo-labels each round from an exponential moving average (EMA) teacher~$\teacher_{t}$ that evolves with the global model, defined as $\teacher_{t} = \lambda\,\teacher_{t-1} + (1-\lambda)\,\student_t$, where $\lambda$ is the EMA decay rate.
Along the \emph{server training} axis, Static PL performs no server-side updates; FedNST computes a weighted average of gradients from the full server update (gradient descent) and client pseudo-gradients; Rao~\etal augment each FL round with supervised gradients from held-out labeled server clients; and Diao~\etal alternate FL aggregation rounds with a server-side supervised pass on a single batch.\footnote{FL update and server update are merged together to update the student model $\student$ once in FedNST and Rao~\etal}
We provide a per-method description of how FedNST and \citet{rao2023fl_self} instantiate these two axes in \Cref{app:baselines}.

Note that \citet{diao2022semifl} validate their proposed method on image classification tasks (not ASR tasks). 
However, as it is one of the most effective methods in SSFL for image classification and less bound to image classification tasks\footnote{\citet{diao2022semifl} employ FixMatch~\citep{li2024safixmatch} but it is not straightforward to apply such algorithms to ASR tasks due to variable-length sequences and non-unique alignment of CTC.}, we modify and employ it as a baseline for comparison. 
Interestingly, \Cref{fig:pl-source-basics-baselines} shows that \citet{diao2022semifl} for ASR significantly outperform other baselines: Static PL, FedNST and \citet{rao2023fl_self} when $\serverD:$ LS100 and $\clientD$: LS860 \vTwoEdits{(LibriSpeech subsets; \Cref{subsec:datasets})}.
Therefore, we focus our analysis on comparison with the tailored \citet{diao2022semifl} for ASR.

\paragraph{Global vs. online teacher.}
Let $\hat{f}(\cdot;\theta)$ denote the decoding function parameterized by $\theta$ generating pseudo-labels.
At round $t$, local step $\tau$, on client $i$ and sample $\mathbf{x}_{i,j}$, the pseudo-label $\hat{\mathbf{y}}_{i,j}$ is produced from one of two sources:
\begin{align}
\textbf{Global teacher:}\quad & \hat{\mathbf{y}}_{i,j} = \hat{f}\!\left(\mathbf{x}_{i,j};\ \teacher_t\right), \label{eq:global-teacher}\\
\textbf{Online teacher:}\quad & \hat{\mathbf{y}}_{i,j} = \hat{f}\!\left(\mathbf{x}_{i,j};\ \student_{t,\tau}\right), \label{eq:online-teacher}
\end{align}
where $\teacher_t$ is \vOneEdits{the teacher model, maintained as an exponential moving average of the broadcast global model across rounds ($\teacher_t = \lambda\,\teacher_{t-1} + (1-\lambda)\,\student_t$, $\lambda=0.99$; \Cref{app:fl})}, held fixed across all local steps within the round, and $\student_{t,\tau}$ is the client's local parameters at step $\tau \in \{0, 1, \ldots, K-1\}$, which evolve during local training.

The two teachers therefore differ only when clients take more than one local step: with a single step ($K=1$) the local model has not yet moved, so $\student_{t,\tau}=\teacher_t$ and every client is labeled by the same global model.
Multiple local steps are what let the online teacher adapt to each client within the round, but they trade off against stability---too many steps let the local model drift and its pseudo-labels grow stale or inconsistent, which we study in \Cref{subsec:ablations}.
\vTwoEdits{More fundamentally, multiple local steps are standard in FL because communication is the bottleneck---each round pays network latency and a synchronous aggregation barrier, whereas local computation is cheap---so more local work per round reduces the number of communication rounds \citep{mcmahan2017fl}, and can even accelerate convergence \citep{mishchenko2022proxskip}.}
\Cref{tab:teachers} summarizes these two sources together with the local EMA and transitioning teachers introduced in \Cref{sec:pl-generation}.
The remainder of the paper analyzes this choice---and its combination with server update---across progressively harder domain-shift regimes.

\subsection{Practical Constraints and Scope}
\label{subsec:constraints}

\paragraph{Practical constraints.}
\vOneEdits{We adopt three simplifying constraints that shape our experimental design; they are not intended to characterize any particular production FL system:}
\begin{enumerate}
    \item No labeled data exists on clients; only the server holds a labeled seed dataset.
    \item Clients are \emph{non-persistent} across rounds---a client selected in one round may never be selected again---so strategies that accumulate per-client state (e.g., a local model checkpoint) across rounds are fragile.
    \item Communication overhead must remain minimal: only a single model can be broadcast per round, ruling out approaches that require transmitting separate teacher and student snapshots.
\end{enumerate}
Together, these constraints narrow the design space to methods that differ in how pseudo-labels are generated each round and how the server re-injects the labeled signal into the global model---the two axes of Table~\ref{tab:method_comparison}.

\paragraph{Scope.}
Several techniques have been shown to improve pseudo-label quality in centralized semi-supervised ASR but are orthogonal to the two axes studied here.
Language-model-guided decoding during pseudo-label generation---such as LM beam search~\citep{xu2020ipl} or LM fusion~\citep{park2020nst,zhang2020pushinglimits}---can improve transcription quality regardless of whether the pseudo-label source is a static seed or an evolving EMA teacher.
Similarly, data filtering based on confidence or uncertainty, as employed by \citet{rao2023fl_self}, can be layered on top of any pseudo-label source and server training strategy.
Because these mechanisms can be independently combined with the choices we study, we hold them fixed (no LM, no filtering) throughout our experiments to isolate the effect of the pseudo-label source and server training strategy.

\section{Experimental Setup}
\label{sec:setup}

\vOneEdits{All configurations described below and in \Cref{app:setup} reflect choices for our public-dataset simulations.}

\subsection{Datasets and Domain-Shift Pairs}
\label{subsec:datasets}

We evaluate methods on four popular English ASR corpora spanning a range of speaking styles and recording conditions: LibriSpeech (LS, read audiobooks), TED-LIUM (Ted, TED talks), Common Voice (CV, crowd-sourced read speech) and Fisher (telephone conversations).
\Cref{tab:datasets} summarizes each corpus (hours, number of speakers, and utterance durations).
Throughout, LS$x$ denotes an $x$-hour labeled subset of LS, and CV10 and CV90 a $10\%$/$90\%$ split of CV; subsets of the same corpus (e.g., LS100 and LS860) are drawn from the same distribution and use disjoint speakers, differing only in size.

Experiments are conducted on server (source) $\rightarrow$ client (target) pairs, for example, LS100 $\rightarrow$ LS860 denotes $\serverD:$ LS100 and $\clientD:$ LS860.
These pairs range from in-domain (server and client drawn from the same corpus) to cross-domain (different corpora).
As one possible way to measure this domain shift, an embedding-based similarity measure (\Cref{app:domain-shift}) cleanly separates the in-domain pairs from the cross-domain ones, and can serve as a quantitative proxy for the server--client overlap.

For client datasets, we split datasets into small per speaker datasets, treating each speaker as a \vOneEdits{simulated} client for federated learning. 
For server datasets, we do not split them and sample batches uniformly randomly after filtering out audio samples where their duration is too long, e.g., $30$ sec.
Across every pair, the server (seed) and client speaker sets are disjoint---no speaker appears in both---so clients never observe the labeled seed. \vTwoEdits{In every pair the server holds a small amount of labeled seed data and the clients hold a larger, disjoint pool of unlabeled data. For the in-domain pairs the seed and client data are non-overlapping splits of a single corpus (LS100 $\rightarrow$ LS860: $100$\,h of labeled seed data vs. $860$\,h of unlabeled client data; CV10 $\rightarrow$ CV90: $10\%$ labeled seed data vs. $90\%$ unlabeled data distributed across clients); for the cross-domain pairs a small labeled seed corpus is paired with a larger unlabeled client corpus from a different domain (e.g., $100$\,h of LS against ${\sim}1{,}593$\,h of CV). Only the server data is labeled.}

For every pair, we report WER on both the server and client datasets, since a strategy that improves target WER at the cost of significant regression on server data is not a practical win.
All results are reported on the dev sets, as they form part of our analysis and hyperparameter tuning, except for the final comprehensive comparison in \Cref{tab:min-wer-global-vs-online}, which we report on the test sets.
Per-dataset statistics and train, dev, and test splits are deferred to \Cref{app:datasets}.

\subsection{Model Architecture and Training Configuration}
\label{subsec:model}

\paragraph{Model.} All experiments use a Transformer~\citep{vaswani2017transformer} encoder trained with CTC loss~\citep{graves2006ctc} on 80-dim log-mel filterbank features, a character-level tokenizer, and greedy decoding.
Despite its simplicity, greedy decoding is shown to be as effective as more costly decoding algorithms such as beam search for semi-supervised learning for ASR tasks~\citep{likhomanenko2021slimipl}.

As stated in \cref{subsec:constraints}, we deliberately omit both language-model-guided decoding and confidence-based data filtering: these are orthogonal to our two axes of study---pseudo-label generation and server training---and either could be added on top of any configuration we evaluate.

\paragraph{Training.}
For training, audio feature inputs are augmented with SpecAugment~\citep{park2019specaugment} although we use clean features without augmentation to generate pseudo-labels.
We use a simple SGD~\citep{robbins1951sgd} without learning rate scheduler for local updates on the client side and LAMB~\citep{you2020lamb} on the server side for both FL updates and server updates on the labeled server data $\serverD$.
Note that we deliberately choose a layer-wise adaptive optimizer, e.g., LAMB, following \citet{azam2023fl4asr}.

The seed model---trained on the server's labeled data $\serverD$---initializes both the student $\student$ and the EMA teacher $\teacher$ used for pseudo-labeling.
In addition, we reduce both dropout and layer dropout rate from $0.3$ (for seed model training) to $0.1$ during SSFL to increase the model capacity required to learn from large unlabeled data, following \citet{likhomanenko2021slimipl}.
More detailed architecture and training hyperparameters are provided in \Cref{app:model}.

\subsection{Semi-Supervised Federated Learning Configuration}
\label{subsec:fl-config}

We aggregate client updates with FedAvg~\citep{mcmahan2017fl}.
Each sampled client $i \in \clients_t$ trains locally from the broadcast global model $\student_t$ and returns a pseudo-gradient $g_t^{i} = \student_t - \student_t^{i}$, where $\student_t^{i}$ are its locally updated parameters.
FedAvg takes their unweighted mean, which the server optimizer then applies to obtain the next global model:
\begin{equation}
\bar{g}_t = \frac{1}{|\clients_t|} \sum_{i \in \clients_t} g_t^{i}, \qquad \student_{t+1} = \textsc{ServerOpt}\!\left(\student_t,\, \bar{g}_t\right),
\label{eq:fedavg}
\end{equation}
so every sampled client contributes equally regardless of how much data it holds.
The server optimizer $\textsc{ServerOpt}$ is LAMB (\Cref{subsec:model}).

The server applies a supervised update on $\serverD$ each round with probability $p$ (\emph{server training probability}) as specified in line 12 of \Cref{alg:fl_slimipl}. 
For example, $p = 0.5$ corresponds in expectation to alternating training---a supervised update every other round on average---while $p = 1$ applies server training every round and $p = 0$ disables it entirely. 
This server update on $\serverD$ is crucial to reduce instability of SSFL as will be shown in \Cref{sec:pl-generation}.

To reflect the slow training process of FL, we constrain FL steps to be no more than $6$k.
Also, due to large discrepancy in audio duration per sample, instead of running local updates to the fixed number of epochs, we run it to the fixed number of local steps: $160$ for $\clientD:$ LS, and $20$ for the other datasets unless specified otherwise.
Other hyperparameters such as communication rounds, cohort size, local steps, EMA decay, and the exact batch-size values are listed in \Cref{app:fl}.

\begin{figure*}[t]
\centering
\begin{subfigure}[t]{0.48\textwidth}
    \centering
    \includegraphics[width=\linewidth]{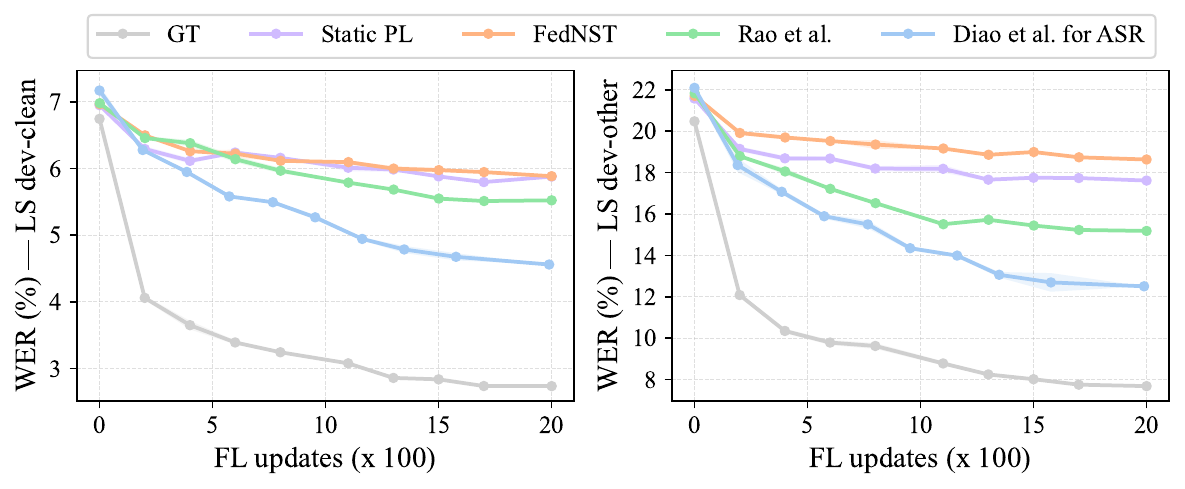}
    \caption{Baseline comparison.}
    \label{fig:pl-source-basics-baselines}
\end{subfigure}
\hfill
\begin{subfigure}[t]{0.48\textwidth}
    \centering
    \begin{minipage}[b]{0.49\linewidth}
        \centering
        \includegraphics[width=\linewidth]{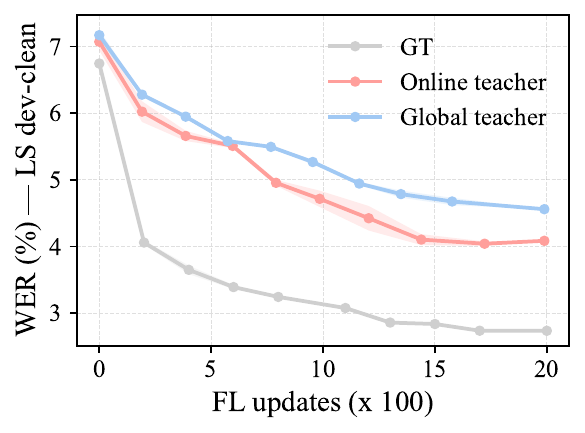}
    \end{minipage}%
    \hfill
    \begin{minipage}[b]{0.49\linewidth}
        \centering
        \includegraphics[width=\linewidth]{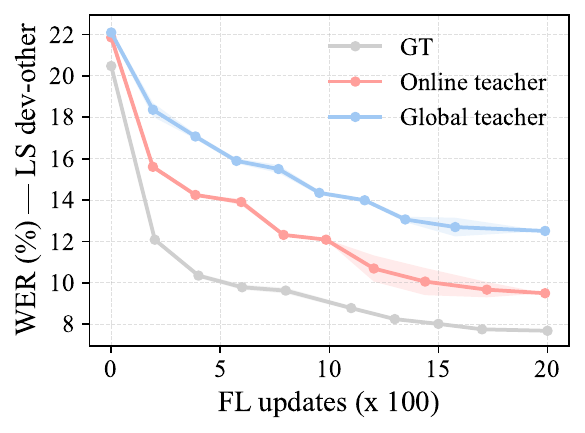}
    \end{minipage}
    \caption{LS dev-clean and dev-other.}
    \label{fig:pl-source-basics-clean}
    \label{fig:pl-source-basics-other}
\end{subfigure}
\hfill
\caption{Panel (a): comparison of the online teacher against existing SSFL baselines---Static PL, FedNST, Rao et al., and Diao et al. for ASR---on LS dev-clean and dev-other. Panel (b): online-teacher advantage holds on both LS dev-clean (left) and dev-other (right) test sets, with online consistently outperforming global throughout training.}
\label{fig:pl-source-basics}
\end{figure*}

\section{Pseudo-Label Generation}
\label{sec:pl-generation}

\vTwoEdits{This section compares the performance of four pseudo-label teachers---global, online, local EMA, and transitioning; their definitions are summarized in \Cref{tab:teachers}.}

\subsection{Global vs. Online Teacher}
\label{subsec:global-vs-online}

Prior SSFL work generates pseudo-labels from either a fixed initial teacher $\teacher_0$~\citep{mehmood2022fednst} or a global EMA teacher $\teacher_t$ updated each round~\citep{diao2022semifl,rao2023fl_self}; none consider an online teacher whose state evolves within the round, reflecting a prevailing assumption---made explicit by \citet{diao2022semifl,zhao2024fedswitch}---that local updates would drift from the global model and destabilize training.
We revisit this choice first in the in-domain regime (LS100 $\rightarrow$ LS860), where the seed and target share the same distribution and any local drift moves the model along in-distribution directions.
All experiments in this section use SpecAugment on client audio and a periodic server-side supervised pass on $\serverD$ determined by server training probability $p=0.5$. We defer analysis of the server training frequency to \Cref{subsec:online-requirements}.

The right panel of \Cref{fig:pl-source-basics-other} plots target-domain WER over FL rounds for both global and online teachers (Eqs.~\ref{eq:global-teacher},\,\ref{eq:online-teacher}; defined in \Cref{subsec:problem-setting}).
The online teacher reaches lower WER than the global teacher from the first few FL rounds onward, and the gap persists rather than vanishing as training proceeds ($\Delta\text{WER} = 2.88$ in the final round).
This improvement is not at the cost of regression on the source domain. WER on dev-clean (left in \Cref{fig:pl-source-basics-clean}) stays within $\Delta\text{WER} = 0.5$ of the global teacher.

The mechanism is direct: the client's local updates refine the model along useful in-distribution directions, so the online teacher's pseudo-labels reflect what the student has just learned on the current client's audio, while the global teacher's pseudo-labels remain stale for the rest of the round.
Unlike FedSwitch~\citep{zhao2024fedswitch}---which switches adaptively between a global teacher and a teacher-student EMA but relies on per-client state---the unconditional online teacher already suffices in-domain under non-persistent clients.

\begin{figure*}[t]
\centering
\begin{subfigure}[t]{0.32\textwidth}
    \centering
    \includegraphics[width=\linewidth]{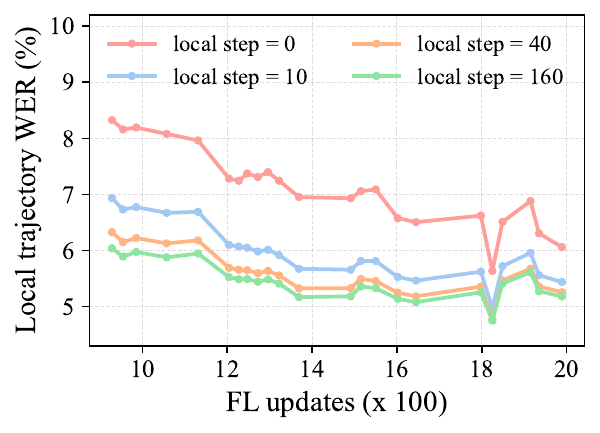}
    \caption{Intermediate local steps.}
    \label{fig:online-validate-localstep}
\end{subfigure}
\hfill
\begin{subfigure}[t]{0.32\textwidth}
    \centering
    \includegraphics[width=\linewidth]{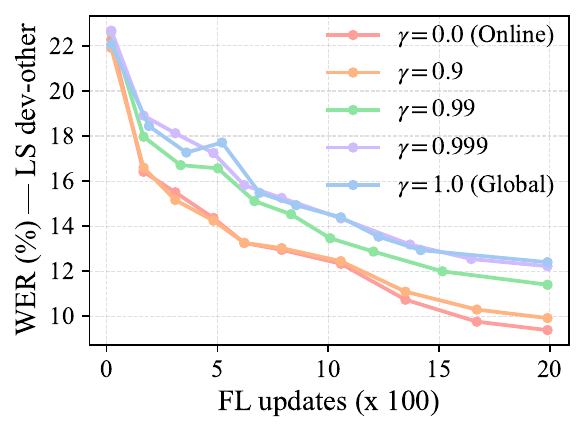}
    \caption{Local EMA decay.}
    \label{fig:pl-source-basics-spectrum}
\end{subfigure}
\hfill
\begin{subfigure}[t]{0.32\textwidth}
    \centering
    \includegraphics[width=\linewidth]{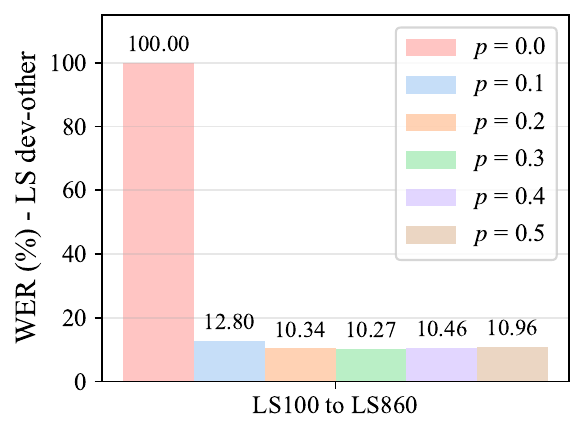}
    \caption{Server training probability.}
    \label{fig:pl-source-basics-failure}
\end{subfigure}
\caption{Mechanism validations on LS100 $\rightarrow$ LS860. (a)~Local-model WER at intermediate local steps $\{0, 10, 40, 160\}$. Step 0 corresponds to the global teacher (no local updates yet) and the last step to the online teacher; WER drops monotonically as local training proceeds, validating that within-round adaptation drives the online teacher's gain. (b)~WER over FL rounds for the global, online, and local-EMA teachers across EMA decay rates (LS dev-other). Local EMA underperforms online at every decay rate, suggesting the latest adaptation step matters more than smoothed local state. (c)~Bar chart comparing online-teacher WER across server training probabilities $p$. Sufficient probability ($p \ge 0.2$) keeps the online teacher stable. Lower (or zero) probability causes divergence as the global model drifts from the labeled distribution.}
\label{fig:online-validate}
\end{figure*}

\paragraph{Local adaptation drives the gain.}
We investigate why the online teacher performs better than the global teacher by tracking how a client's predictions improve over its local training within a single FL round.
At local SGD steps $\{0, 10, 40, 160\}$---where step $0$ is the server parameters at the start of the round before any local update---we take each client's local model and decode a shared client validation set, computing the WER against its references.
The local-trajectory WER at step $s$ is the mean of these per-client WERs across the cohort.
Step $0$ therefore corresponds to the global teacher and the final step to the online teacher.
\Cref{fig:online-validate-localstep} shows that this WER drops monotonically as local training proceeds ($0$ step $\rightarrow$ $160$ step), so the local model adapts to the client distribution within the round.
This adaptation is exactly why the online teacher outperforms the global teacher: it generates pseudo-labels from the adapted local model, whereas the global teacher uses the unadapted server model from the start of the round.

\paragraph{The latest local step matters most.}
We investigate further if interpolation between global and online teachers improves WER compared to the online teacher. 
To this end, we consider a more generic teacher definition based on EMA decay, which we call the local EMA teacher. 
More formally, the local EMA teacher is updated at local step $\tau$ as: $\local_{t,\tau} = \gamma \local_{t, \tau-1} + (1 - \gamma) \student_{t, \tau}$. 
When $\gamma = 0.0$, it is equivalent to the online teacher whereas when $\gamma = 1.0$, it is equivalent to the global teacher.

\Cref{fig:pl-source-basics-spectrum} provides the comparison between $\gamma = \{0.0, 0.9, 0.99, 0.999, 1.0 \}$.
It demonstrates that decreasing $\gamma$ (closer to the online teacher) gradually improves WER, showing that fast adaptivity of the teacher seems to be an important factor.

\subsection{Online Teacher Requires Frequent Server Training}
\label{subsec:online-requirements}

The online teacher's stability hinges on \emph{frequent} server updates: without periodic re-injection of supervised signal from the labeled server dataset $\serverD$, the global model drifts from the labeled distribution across rounds, pseudo-label quality degrades cumulatively, and the online teacher eventually diverges.

\Cref{fig:pl-source-basics-failure} sweeps the server training probability $p$ (defined in \Cref{subsec:fl-config}) on LS100 $\rightarrow$ LS860 across $p \in \{0.0, 0.1, \ldots, 0.5\}$: at $p=0$ training diverges, at $p=0.1$ it remains unstable, resulting in WER $= 12.8$.
However, at $p \ge 0.2$ the online teacher converges to the $10$–$11$ range in WER, with $p=0.3$ giving the best converged WER of $10.27$.
A moderately frequent server update thus suffices: once pseudo-label drift is suppressed, any $p \ge 0.2$ works with no significant additional benefit from raising $p$ further.
Therefore, we fix $p = 0.5$ throughout the paper unless stated otherwise.
This sensitivity to $p$ motivates the deeper investigation of server update strategy in \Cref{sec:server-update}.

\subsection{Cross-Domain Robustness}
\label{subsec:cross-domain-fixed}

To check whether the in-domain finding of \Cref{subsec:global-vs-online} extends beyond LS100 $\rightarrow$ LS860, we evaluate the same global vs. online comparison on two cross-domain targets: LS100 $\rightarrow$ CV (read audiobooks $\rightarrow$ crowd-sourced read speech) and LS100 $\rightarrow$ Ted (read audiobooks $\rightarrow$ prepared talks).

\begin{figure*}[t]
\centering
\begin{subfigure}[t]{0.48\textwidth}
    \centering
    \begin{minipage}[b]{0.49\linewidth}
        \centering
        \includegraphics[width=\linewidth]{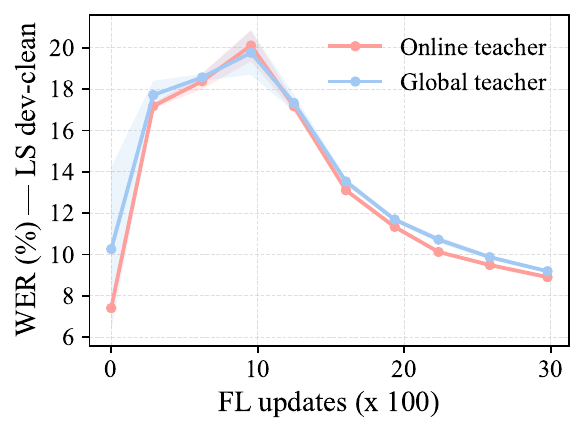}
    \end{minipage}%
    \hfill
    \begin{minipage}[b]{0.49\linewidth}
        \centering
        \includegraphics[width=\linewidth]{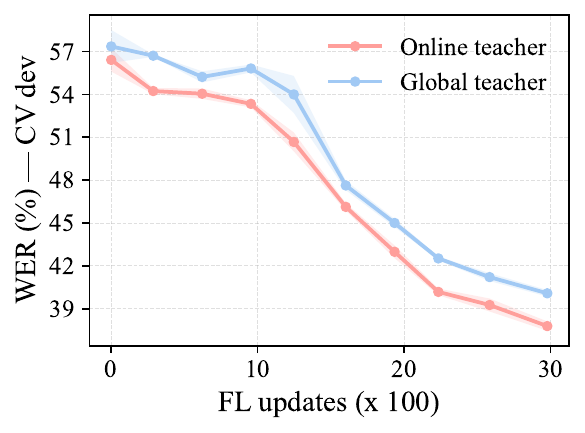}
    \end{minipage}
    \caption{LS100 $\rightarrow$ CV: dev-clean and en-dev.}
    \label{fig:cross-domain-fixed-cv-clean}
    \label{fig:cross-domain-fixed-cv}
\end{subfigure}
\hfill
\begin{subfigure}[t]{0.48\textwidth}
    \centering
    \begin{minipage}[b]{0.49\linewidth}
        \centering
        \includegraphics[width=\linewidth]{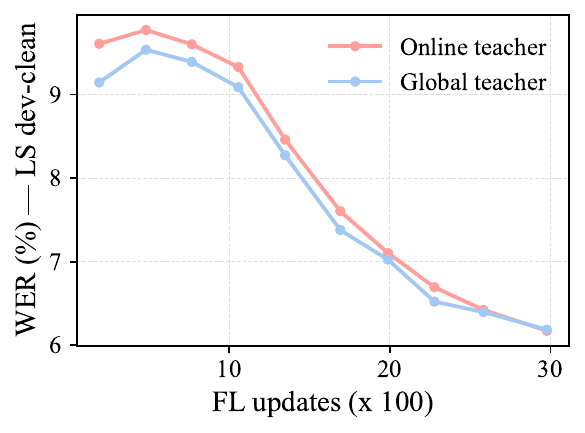}
    \end{minipage}%
    \hfill
    \begin{minipage}[b]{0.49\linewidth}
        \centering
        \includegraphics[width=\linewidth]{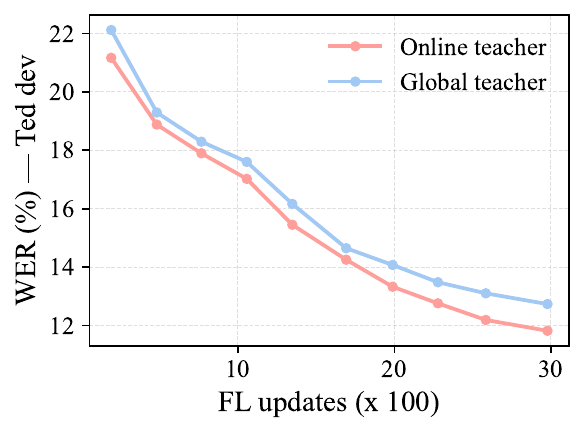}
    \end{minipage}
    \caption{LS100 $\rightarrow$ Ted: dev-clean and ted-dev.}
    \label{fig:cross-domain-fixed-ted-clean}
    \label{fig:cross-domain-fixed-ted}
\end{subfigure}
\caption{Online vs. global teacher at fixed seed (LS100) on two cross-domain targets. Panel (a): LS100 $\rightarrow$ CV (read audiobooks $\rightarrow$ crowd-sourced read speech), evaluated on the source-domain (LS dev-clean, left) and target-domain (CV en-dev, right) test sets. Panel (b): LS100 $\rightarrow$ Ted (read audiobooks $\rightarrow$ prepared talks), evaluated on the source-domain (LS dev-clean, left) and target-domain (Ted ted-dev, right) test sets. Online continues to outperform global on both pairs, confirming that the in-domain finding of \Cref{subsec:global-vs-online} extends to cross-domain shifts at fixed seed.}
\label{fig:cross-domain-fixed}
\end{figure*}

\Cref{fig:cross-domain-fixed} shows that the online teacher continues to outperform the global teacher on both pairs of server and client datasets. 
On the target validation sets in the right figures of \Cref{fig:cross-domain-fixed}(a)(b), global vs. online show WER on CV-dev: $39.9$ vs. $37.7$ and WER on Ted-dev: $12.6$ vs. $11.6$.
Note that this improvement is not at the cost of regression on the source data as shown in the left figures.
The advantage is therefore not specific to the in-domain regime---it persists across the kinds of cross-domain shifts that arise when the server's labeled seed and the clients' unlabeled data are drawn from different speaking styles and recording conditions.

\subsection{Seed Strength and Transitioning Teacher}
\label{subsec:seed-strength}

\paragraph{Seed strength.}
\vOneEdits{Stronger seeds are intended to approximate more mature hypothetical deployment settings, where the labeled server dataset would have accumulated over time.} 
A seed-strength sweep both tests how the benefit of online teacher generalizes to less-favorable conditions for online, and characterizes the regime where online teacher's local adaptation still pays off.

\begin{figure*}[t]
\centering
\includegraphics[width=0.7\linewidth]{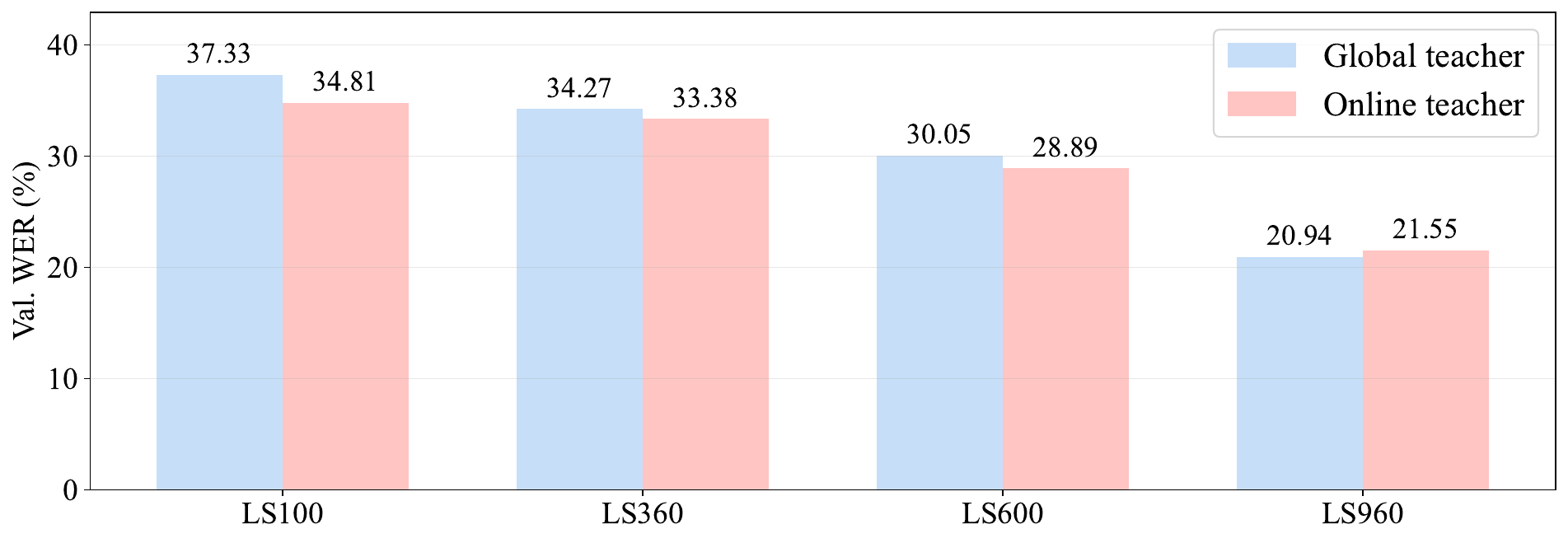}
\caption{Best WER across seed strengths (LS100/360/600/960 $\rightarrow$ CV) for the global, online, and transitioning teacher. The online teacher's advantage over global narrows as the seed grows stronger; the transitioning teacher (global $\rightarrow$ online at round $r$) recovers the gap and matches or beats the better of the two pure strategies at every seed strength.}
\label{fig:domain-shift-transition}
\vspace{-2mm}
\end{figure*}

\Cref{fig:domain-shift-transition} shows that as the seed grows stronger (LS100, 360, 600, 960 $\rightarrow$ CV), the gap between online and global narrows further: $\Delta\text{WER}$ shrinks from $2.52$ at LS100 to $0.89$ at LS360 ($34.27$ vs. $33.38$), $1.16$ at LS600 ($30.05$ vs. $28.89$), and $-0.61$ at LS960 ($20.94$ vs. $21.55$). 
A strong seed already produces good pseudo-labels, leaving less room for the online teacher's local adaptation to improve them.
With the strongest seed (LS960 $\rightarrow$ CV), the online teacher can even underperform the global teacher in early rounds, before the model has had time to adapt to the target domain.
In those early rounds, the strong seed's global predictions on the unseen target distribution still beat the online teacher's, which relies on a model that has barely begun to adapt.

\begin{figure*}[t]
\centering
\begin{subfigure}[t]{0.48\textwidth}
    \centering
    \begin{minipage}[b]{0.49\linewidth}
        \centering
        \includegraphics[width=\linewidth]{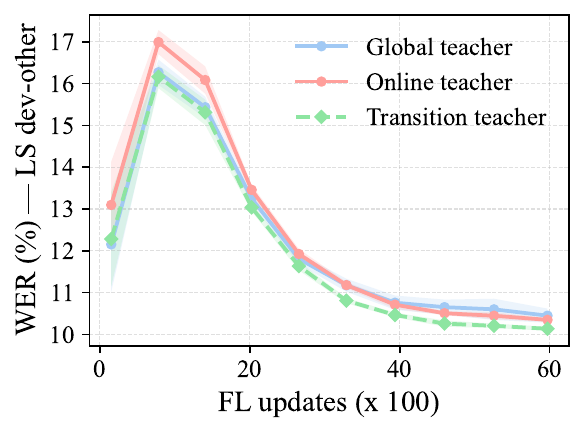}
    \end{minipage}%
    \hfill
    \begin{minipage}[b]{0.49\linewidth}
        \centering
        \includegraphics[width=\linewidth]{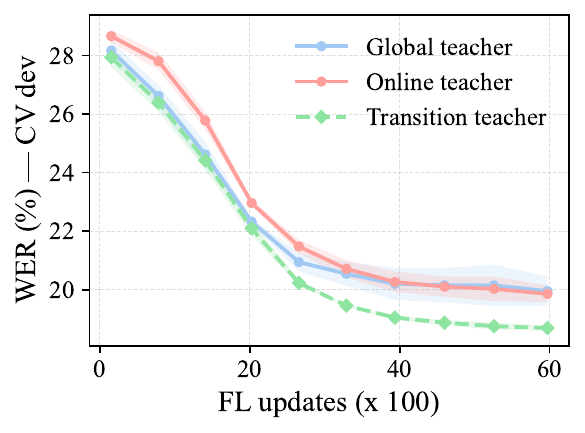}
    \end{minipage}
    \caption{LS960 $\rightarrow$ CV: dev-other and en-dev.}
\end{subfigure}
\begin{subfigure}[t]{0.24\textwidth}
    \centering
    \includegraphics[width=\linewidth]{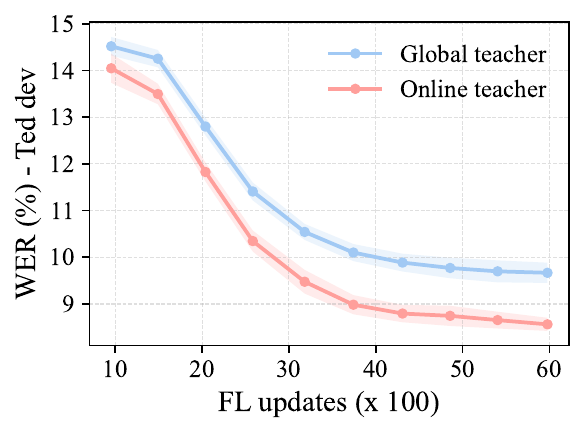}
    \caption{LS $\rightarrow$ Ted.}
    \label{fig:online-breaks-ls-ted}
\end{subfigure}
\begin{subfigure}[t]{0.24\textwidth}
    \centering
    \includegraphics[width=\linewidth]{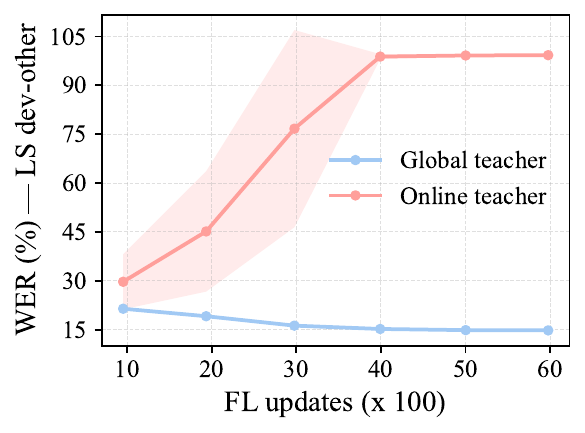}
    \caption{Ted $\rightarrow$ LS.}
    \label{fig:online-breaks-ted-ls}
\end{subfigure}
\caption{Panel (a): training curves on LS960 $\rightarrow$ CV for global, online, and transitioning teachers on the source-domain (LS dev-other, left) and target-domain (CV en-dev, right) test sets: at this strong seed, online slightly underperforms global, while the transitioning teacher (global $\rightarrow$ online at round $r$) improves over both pure strategies on the target. Panels (b,c): direction-asymmetric breakage of the online teacher: LS $\rightarrow$ Ted converges with online outperforming global, whereas Ted $\rightarrow$ LS diverges rapidly. The asymmetry motivates the investigation in \Cref{subsec:why-breaks-server}.}
\label{fig:transition-curves}
\label{fig:online-breaks}
\end{figure*}

\paragraph{Transitioning teacher.}
This early-round disadvantage motivates the \emph{transitioning teacher}, which uses the global teacher for the first $r$ rounds and switches to the online teacher thereafter.
The right panel of \Cref{fig:transition-curves}(a) (CV en-dev, target-domain) shows that the transitioning teacher is almost identical until $r=2$k and switches to the online teacher afterwards, and this transition provides significant improvement over the global teacher by $\Delta\text{WER} = 1.20$ (global: $20.9$ vs. transition: $19.7$).

As stated before, this is because in early rounds the model is far from the target distribution, so the online teacher would produce unreliable pseudo-labels while the global teacher provides stable ones.
Once the model has adapted enough to the target domain, the online teacher's local adaptivity becomes more beneficial.
The optimal crossover round $r^*$ depends on seed strength and domain gap.
For example, with a weaker seed model (LS100 $\rightarrow$ CV), this transition is never beneficial, i.e., $r^* = 0$.
Also, when there is little to no domain gap, e.g., LS100 $\rightarrow$ LS860, the transition is again not beneficial.

\begin{takeaways}
\begin{itemize}[leftmargin=*, itemsep=2pt, topsep=2pt]
    \item Contrary to the common assumption that a per-client \emph{online} teacher is too unstable to use, it matches or beats the broadcast \emph{global} teacher both in-domain and under domain shift.
    \item The online teacher's advantage comes from its within-round adaptation to each client's distribution, but the same adaptivity makes it prone to instability, so it holds only when paired with frequent enough server training on the seed data.
    \item The online teacher's advantage over the global teacher shrinks as the seed grows stronger ($\Delta\text{WER} = 2.52$ at LS100 $\rightarrow$ CV down to $-0.61$ at LS960 $\rightarrow$ CV), but a \emph{transitioning} teacher (global $\rightarrow$ online at round $r$) recovers the gap and matches or beats both at every seed strength.
\end{itemize}
\end{takeaways}

\section{Server-Update Strategy}
\label{sec:server-update}

\Cref{sec:pl-generation} established that the online teacher matches or beats the global teacher across both in-domain and cross-domain settings, including significant domain shift (e.g., LS $\rightarrow$ CV).
However, the online teacher does not always work. This section analyzes its failure modes and how to mitigate them.

\begin{figure*}[t]
\centering
\begin{subfigure}[t]{0.32\textwidth}
    \centering
    \includegraphics[width=\linewidth]{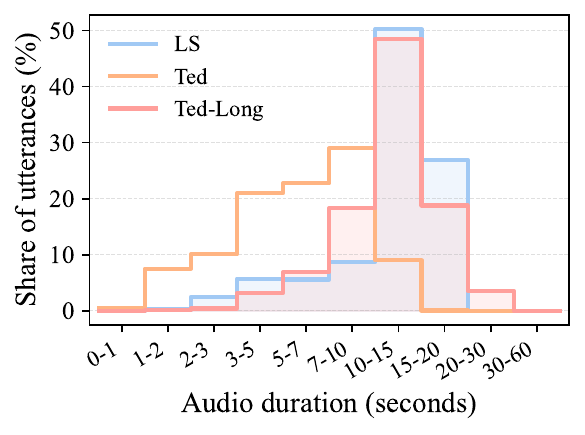}
    \caption{Audio duration distribution.}
    \label{fig:duration-histogram}
\end{subfigure}
\hfill
\begin{subfigure}[t]{0.66\textwidth}
    \centering
    \begin{minipage}[b]{0.49\linewidth}
        \centering
        \includegraphics[width=\linewidth]{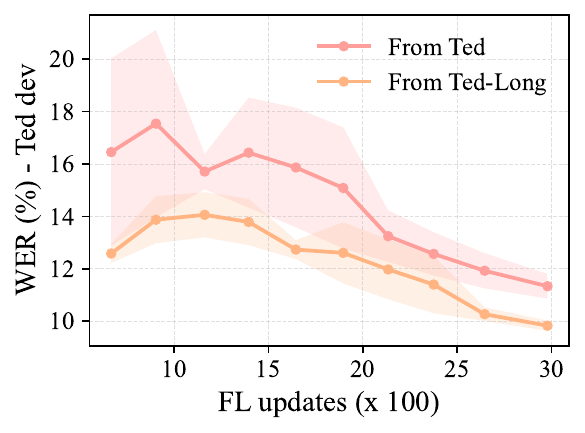}
    \end{minipage}%
    \hfill
    \begin{minipage}[b]{0.49\linewidth}
        \centering
        \includegraphics[width=\linewidth]{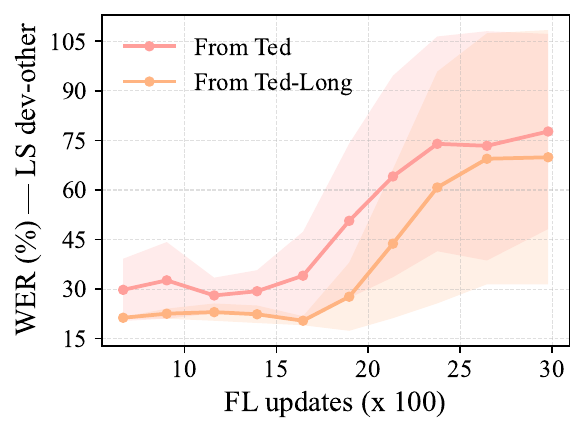}
    \end{minipage}
    \caption{Ted/Ted-Long $\rightarrow$ LS.}
    \label{fig:ted-tedlong-compare}
\end{subfigure}
\caption{Audio duration does not explain the asymmetric breakage. (a)~Ted-Long matches LS in audio duration distribution, while original Ted is markedly shorter. (b)~When the seed is trained on Ted-Long instead of Ted, the Ted-Long $\rightarrow$ LS direction still diverges similarly to Ted $\rightarrow$ LS---ruling out audio duration as the driver of the asymmetry.}
\label{fig:seed-training-duration}
\end{figure*}

\subsection{Online Teacher Breaks Asymmetrically}
\label{subsec:online-breaks}

\paragraph{Asymmetric breakage.}
We extend the online-vs.-global comparison from \Cref{sec:pl-generation} to two new pairs.
On LS $\rightarrow$ Ted (\Cref{fig:online-breaks-ls-ted}), the online teacher outperforms the global teacher as expected.
On the reverse Ted $\rightarrow$ LS (\Cref{fig:online-breaks-ted-ls}), however, the online teacher diverges while the global teacher converges normally.
The same asymmetric pattern holds for LS $\rightarrow$ CV vs. CV $\rightarrow$ LS.

The asymmetry rules out a simple ``large domain gap = failure'' intuition.
In fact, the larger gap (LS $\rightarrow$ CV) still converges with online beating global, while the smaller gap (Ted $\rightarrow$ LS) diverges.
The cause must lie in the \emph{direction}---specifically, in the properties of the source data used for server training.
We first rule out seed-training-side factors before tracing the failure to the FL-stage server-training step itself. We share these negative results to help readers avoid similar pitfalls.

\subsection{Investigating Instability: Seed Training}
\label{subsec:why-breaks-seed}

The server data participates in training at two stages---seed training before FL and server-training updates during FL---so we first check whether the seed model itself is the source of instability.
There are three aspects of seed training we validate as potential cause of instability of the online teacher in the Ted $\rightarrow$ LS direction: i) audio duration of the seed corpus, ii) SpecAug strength, and iii) overfitting of the seed model.

\paragraph{Audio duration.}
\Cref{fig:duration-histogram} shows that Ted's audio duration distribution (mean: 6.07\,s, median: 5.94\,s) is markedly shorter than LS's (mean: 12.30\,s, median: 13.79\,s).
We hypothesize this gap drives the asymmetric breakage: a seed model trained on Ted's short clips may struggle when further trained on LS's longer clips during FL with pseudo-labels.
Because online-teacher pseudo-labels are noisier than global-teacher ones, the combined effect of pseudo-label noise and longer audio could destabilize the online teacher even where the global teacher remains stable.
To validate this, we construct Ted-Long, a Ted variant that concatenates up to two clips while keeping each example under 40 seconds, yielding a duration distribution (mean: 12.08\,s, median: 11.79\,s) close to LS's.

\Cref{fig:ted-tedlong-compare} shows that Ted-Long $\rightarrow$ LS still diverges similarly to Ted $\rightarrow$ LS, even though the Ted-Long seed performs better on Ted dev. This rules out audio duration as the driver of the instability.

\paragraph{SpecAug strength.}
We hypothesize that strong SpecAug during seed training compounds the instability in the Ted $\rightarrow$ LS direction: aggressive masking on the source corpus can cause improper feature learning, producing significantly wrong predictions on out-of-domain data such as LS.
Because online-teacher pseudo-labels are noisier than global-teacher ones, the combined effect of wrong seed predictions and pseudo-label noise could destabilize the online teacher even where the global teacher remains stable.

\begin{figure*}[t]
\centering
\begin{subfigure}[t]{0.48\textwidth}
    \centering
    \includegraphics[width=\linewidth]{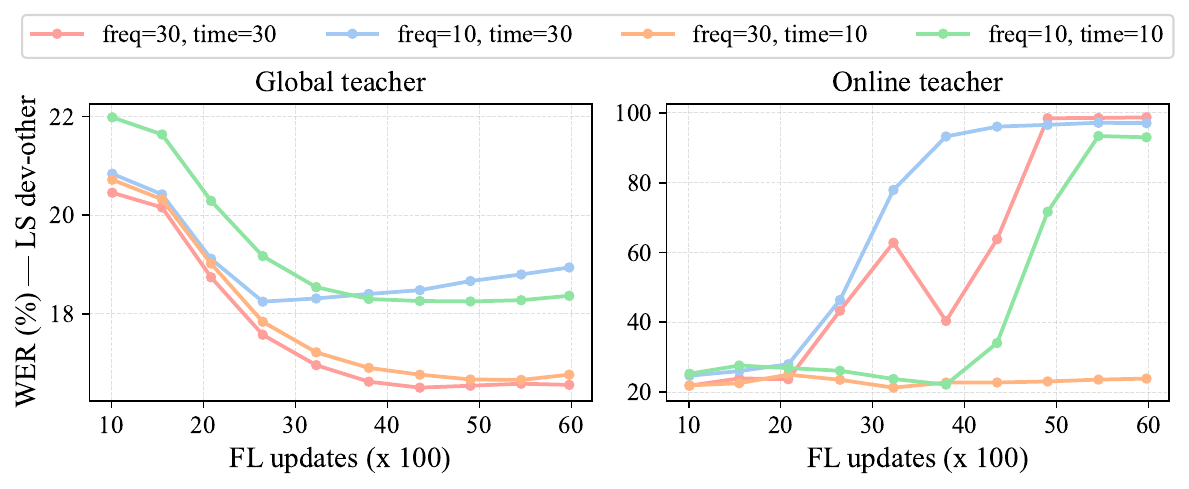}
    \caption{SpecAug strength.}
    \label{fig:seed-specaug}
\end{subfigure}
\hfill
\begin{subfigure}[t]{0.24\textwidth}
    \centering
    \includegraphics[width=\linewidth]{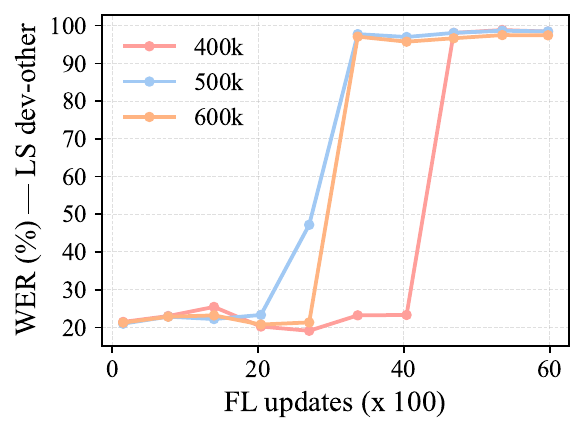}
    \caption{Early seed checkpoint.}
    \label{fig:seed-early-ckpt}
\end{subfigure}
\hfill
\begin{subfigure}[t]{0.24\textwidth}
    \centering
    \includegraphics[width=\linewidth]{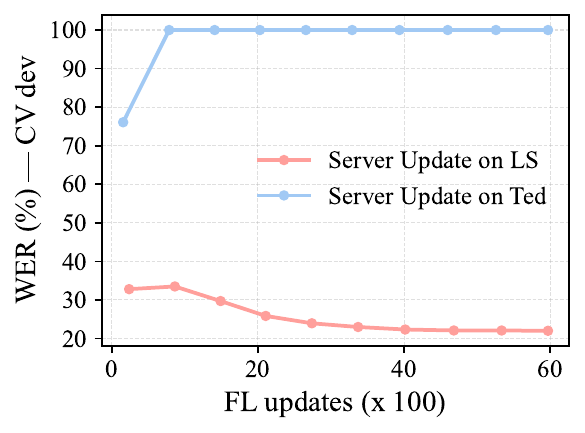}
    \caption{Server update on LS $\rightarrow$ CV.}
    \label{fig:server-update-anchor}
\end{subfigure}
\caption{\emph{Seed-training and server-update factors in the Ted $\rightarrow$ LS instability.} (a)~SpecAug strength sweep over $(W_f, W_t) \in \{(30, 30), (10, 30), (30, 10), (10, 10)\}$ during Ted seed training, evaluated as Ted $\rightarrow$ LS: stronger SpecAug benefits the global teacher, while the online teacher shows no clear pattern. (b)~Seed checkpoint sweep over $\{400\text{k}, 500\text{k}, 600\text{k}\}$ training steps: the 400k checkpoint delays but does not avoid divergence. (c)~With the seed fixed on LS $\rightarrow$ CV, the online teacher diverges when the server update uses Ted but converges when it uses LS.}
\label{fig:seed-overfit}
\end{figure*}

By default, we use the standard SpecAugment recipe: 2 frequency masks of maximum width $W_f = 30$ bins and 10 time masks of maximum width $W_t = 50$ frames (capped at $0.1 \times L$ for utterance length $L$).
To probe this hypothesis, we sweep the maximum mask widths $(W_f, W_t) \in \{(30, 30), (10, 30), (30, 10), (10, 10)\}$ during Ted seed training and evaluate Ted $\rightarrow$ LS in FL with each resulting seed as shown in \Cref{fig:seed-specaug}.

\Cref{fig:seed-specaug} shows an unexpected pattern: stronger SpecAug during seed training is actually beneficial to the global teacher, while the effect on the online teacher is less clear.
A narrower time mask ($W_t = 10$) appears to help: the online teacher does not diverge at $(W_f, W_t) = (30, 10)$.
However, it still diverges at $(10, 10)$, and at $(10, 30)$ divergence is avoided but WER does not improve over the initial seed model.
Overall, this rules out strong SpecAug during seed training as the major driver of the instability.

\paragraph{Overfitting of the seed model.}
Lastly, we check whether overfitting of the seed model to the source corpus contributes to the instability.
We train seed models with varying numbers of steps $\in \{400\text{k}, 500\text{k}, 600\text{k}\}$ on Ted and evaluate Ted $\rightarrow$ LS in FL with each seed.
\Cref{fig:seed-early-ckpt} shows that the seed model trained for fewer steps (400k) diverges later than the default 800k checkpoint, but it does not improve over the initial seed model and still diverges eventually.
This suggests that while overfitting may exacerbate the instability, it is not the root cause of the online teacher's divergence in the Ted $\rightarrow$ LS direction.

In the end, none of these factors explains the asymmetric breakage: in every configuration, Ted $\rightarrow$ LS continues to diverge while LS $\rightarrow$ Ted converges.
The cause must therefore lie in the FL-stage server update, which we examine in \Cref{subsec:why-breaks-server}.

\subsection{Investigating Instability: Server Update}
\label{subsec:why-breaks-server}

An evidence we have for the server update being the source of instability is that the divergence happens depending on which source dataset is used for server update, even when the seed model is fixed.
For example, \Cref{fig:server-update-anchor} shows that when we fix the seed model to be the same checkpoint trained on LS and client data is CV, the online teacher diverges when the server update is performed on Ted but converges when it is performed on LS.
It is worth noting that the global teacher converges in both cases, which further supports the hypothesis that the server update step is the source of instability for the online teacher.

Now, the question is what about the server update step causes the instability and why it only affects the online teacher but not the global teacher.
To investigate this, we analyze the server update step in detail and identify two levers that govern its stability: the SpecAug strength and batch size applied during the server update.

\begin{figure*}[t]
\centering
\begin{subfigure}[t]{0.24\textwidth}
    \centering
    \includegraphics[width=\linewidth]{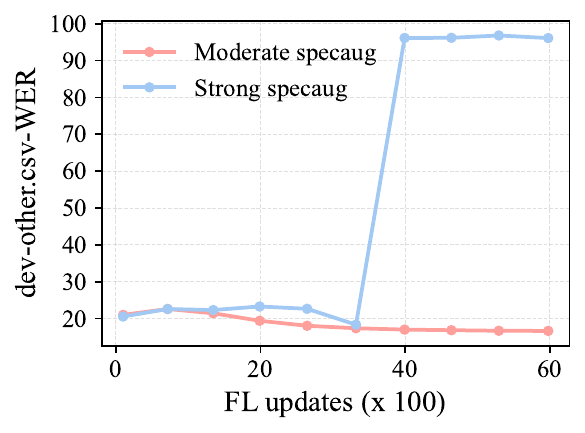}
    \caption{WER on LS dev-other.}
    \label{fig:specaug-sweep-wer}
\end{subfigure}
\hfill
\begin{subfigure}[t]{0.24\textwidth}
    \centering
    \includegraphics[width=\linewidth]{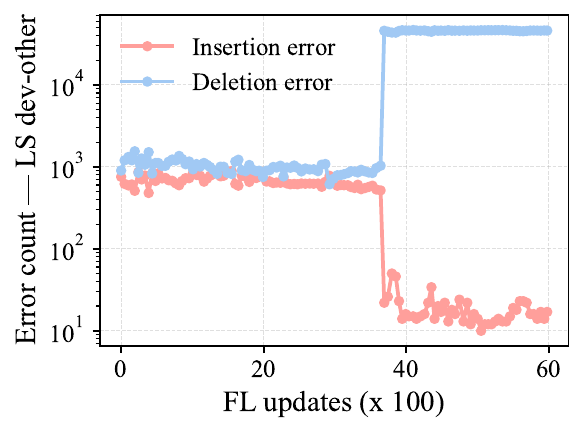}
    \caption{Errors: Strong SpecAug.}
    \label{fig:specaug-sweep-insdel}
\end{subfigure}
\hfill
\begin{subfigure}[t]{0.24\textwidth}
    \centering
    \includegraphics[width=\linewidth]{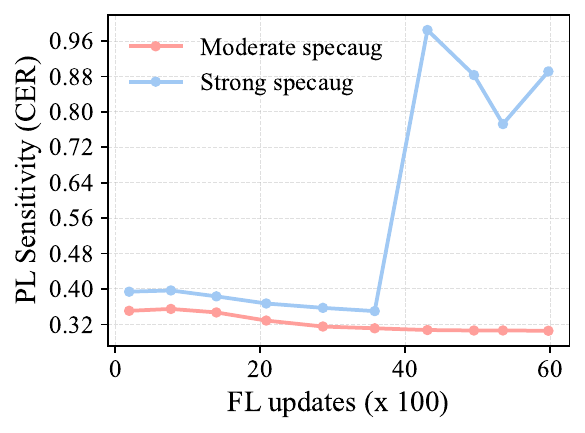}
    \caption{PL Sensitivity (CER).}
    \label{fig:specaug-sweep-plsens}
\end{subfigure}
\hfill
\begin{subfigure}[t]{0.24\textwidth}
    \centering
    \includegraphics[width=\linewidth]{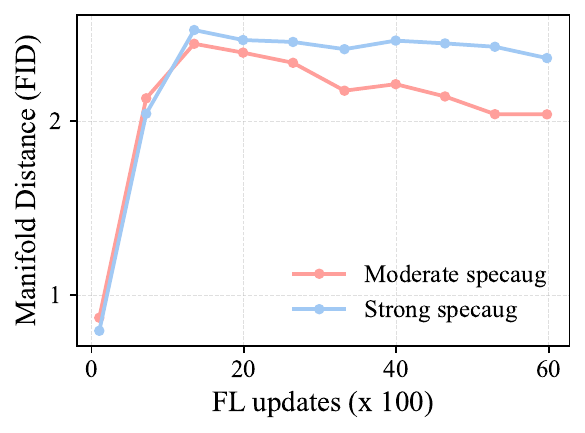}
    \caption{Manifold Dist. (FID).}
    \label{fig:specaug-sweep-manifold}
\end{subfigure}
\caption{Comparison of strong and moderate SpecAug on Ted $\rightarrow$ LS.
(a)~WER on LS dev-other over FL rounds for moderate vs. strong SpecAug.
(b)~Insertion and deletion error rates over FL rounds. Strong SpecAug yields high deletion with near-zero insertion at some FL step due to instability of training.
(c)~PL sensitivity, measured as the FID between pseudo-label distributions on original vs. slightly-perturbed inputs. Strong SpecAug increases sensitivity, evidencing the input-level ill-conditioning of the server pass.
(d)~Manifold distance, measured as the FID between the model's hidden-state distribution and the seed-trained reference. Strong SpecAug pushes the model further off-manifold.}
\label{fig:specaug-sweep}
\end{figure*}

\subsubsection{SpecAug Strength}

\paragraph{Setup.}
As a reminder, the default SpecAug parameters are 2 frequency masks of maximum width $W_f = 30$ bins and 10 time masks of maximum width $W_t = 50$ frames (capped at $0.1 \times L$ for utterance length $L$).
We refer to this default setting as \emph{strong} SpecAug, and we refer to a milder setting of $(W_f, W_t) = (15, 25)$ as \emph{moderate} SpecAug.

\paragraph{WER and error-type signature.}
We compare moderate and strong SpecAug on the server update for Ted $\rightarrow$ LS in \Cref{fig:specaug-sweep-wer}, which shows strong SpecAug diverges while moderate SpecAug converges to a better WER than the seed model.
We further investigate the mechanism of this divergence by analyzing the error types.
\Cref{fig:specaug-sweep-insdel} shows that strong SpecAug yields high deletion error and near-zero insertion error starting from around $3.8$k FL steps.
This error signature implies that after around 3.8k FL steps, the model outputs nothing or very little meaningless transcription.

\paragraph{Pseudo-label sensitivity.}
We conjecture that this is because client model predictions become highly inconsistent across clients under strong SpecAug, which produces noisy pseudo-labels that are dominated by the randomness of mask sampling rather than the content of the input.
This inconsistency is evidenced by the pseudo-label sensitivity to small input perturbations as shown in \Cref{fig:specaug-sweep-plsens}.
We measure ``pseudo-label (PL) sensitivity'' as the average pairwise character error rate (CER) between $k$ independently SpecAug-perturbed decodings of the same utterance, averaged over 256 fixed evaluation utterances.
Lower values indicate more stable predictions under augmentation noise.
We observe that the PL sensitivity suddenly increases around 3.8k FL steps under strong SpecAug, coinciding with the onset of high deletion error.

\paragraph{Manifold drift.}
However, we still do not understand how strong SpecAug leads to this inconsistency.
We hypothesize that strong SpecAug adds strong noise to server updates, which pushes the model off the manifold of the seed data and into a regime where the model's predictions are highly unstable under input perturbations.
To validate this, we measure the distance between the model's hidden-state distribution at the current step and at initialization using the Fr\'{e}chet distance, shown in \Cref{fig:specaug-sweep-manifold}.
Specifically, we pass 512 fixed samples from the server dataset through the model, extract penultimate-layer features (mean-pooled over time), and fit a Gaussian $(\mu_t, \Sigma_t)$ at step $t$. The manifold distance at step $t$ is the Fr\'{e}chet distance to the reference distribution $(\mu_0, \Sigma_0)$ at initialization:
\begin{equation}
\mathrm{FID}_t = \|\mu_0 - \mu_t\|^2 + \mathrm{Tr}\!\left(\Sigma_0 + \Sigma_t - 2(\Sigma_0 \Sigma_t)^{1/2}\right).
\end{equation}
Higher values indicate larger representation drift from the seed-trained reference.
\Cref{fig:specaug-sweep-manifold} shows that the manifold of strong SpecAug drifts much further from the seed model reference than that of moderate SpecAug, supporting the hypothesis that strong SpecAug pushes the model into a regime of unstable predictions.

\begin{figure*}[t]
\centering
\begin{subfigure}[t]{0.32\textwidth}
    \centering
    \includegraphics[width=\linewidth]{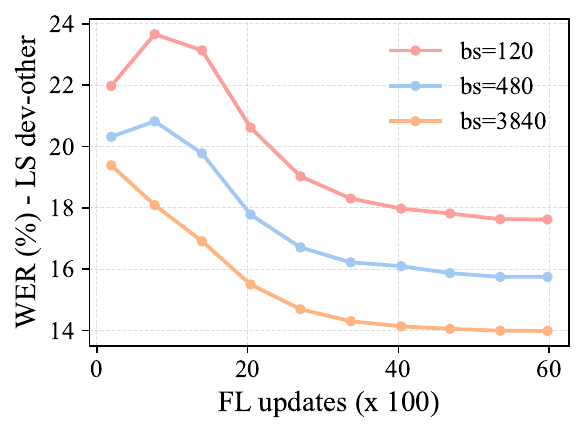}
    \caption{WER on LS dev-other.}
    \label{fig:batch-sweep-insdel}
\end{subfigure}
\hfill
\begin{subfigure}[t]{0.32\textwidth}
    \centering
    \includegraphics[width=\linewidth]{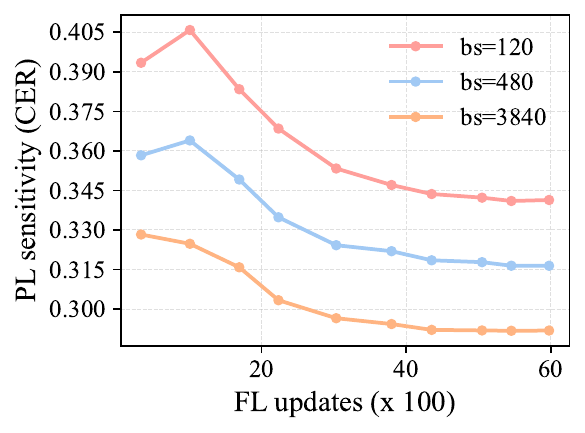}
    \caption{PL Sensitivity (CER).}
    \label{fig:batch-sweep-plsens}
\end{subfigure}
\hfill
\begin{subfigure}[t]{0.32\textwidth}
    \centering
    \includegraphics[width=\linewidth]{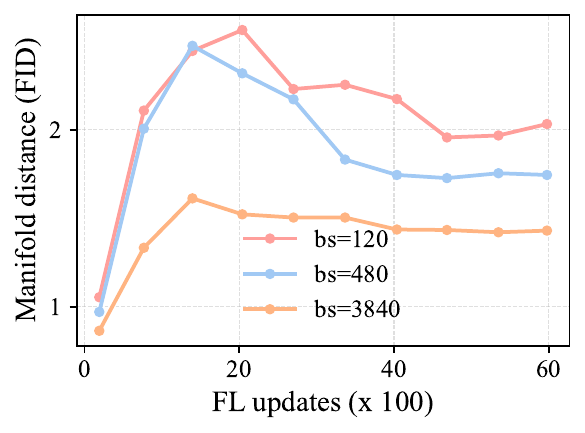}
    \caption{Manifold Distance (FID).}
    \label{fig:batch-sweep-manifold}
\end{subfigure}
\caption{\emph{Batch size sweep at fixed moderate SpecAug on Ted $\rightarrow$ LS} (dev-other).
(a)~WER over FL rounds on LS dev-other. Small batch converges along a noisier trajectory to a higher WER, while larger batches converge to a low, stable WER.
(b)~PL sensitivity, measured as the FID between pseudo-label distributions on original vs. slightly-perturbed inputs. Small batch yields higher sensitivity, evidencing the optimizer-level ill-conditioning of the server pass.
(c)~Manifold distance, measured as the FID between the model's hidden-state distribution and the seed-trained reference. Small batch pushes the model further off-manifold.}
\label{fig:batch-sweep}
\end{figure*}

\subsubsection{Batch Size}
\paragraph{Setup.}
Now that we know strong SpecAug causes server-training instability, we fix SpecAug at moderate and sweep the server-training batch size from small (120) to large (3840).
Throughout the paper, the batch size is measured in seconds of audio rather than number of samples: a batch of 120 means we select samples whose total duration sums to approximately 120 seconds per server-update step. We use duration-based batching---rather than a fixed number of samples---because audio length varies substantially across utterances (typically 2--30 seconds), and this keeps per-step compute consistent.
When batch size is too large to fit in memory, we accumulate gradients over multiple forward passes before performing the optimizer step.
Again, we analyze the effect of batch size on the Ted $\rightarrow$ LS setting.

\paragraph{WER signature.}
Unlike strong SpecAug, small batch does not cause divergence---training converges in all cases (\Cref{fig:batch-sweep-insdel}).
Instead, it converges along a noisier trajectory to a noticeably higher final WER, and increasing the batch progressively recovers stable training, with batch 3840 fully recovering under moderate SpecAug.
This is initially counterintuitive: because we clip the gradient to magnitude 1, a larger batch does not anchor the update with a larger step---it only sharpens the \emph{direction} of the update.
The benefit of large batch is therefore entirely a reduction in gradient-direction noise, which points to an optimizer-level mechanism rather than the input-level one behind strong SpecAug.

\paragraph{Pseudo-label sensitivity.}
As with SpecAug, we probe this instability through PL sensitivity (\Cref{fig:batch-sweep-plsens}).
Small batch yields higher sensitivity to input perturbations, evidencing that the noisy server gradient leaves client predictions less stable---the optimizer-level counterpart of the input-level ill-conditioning caused by strong SpecAug.

\paragraph{Manifold drift.}
The same noise pushes the model off the seed-data manifold.
Using the manifold distance defined above, \Cref{fig:batch-sweep-manifold} shows that small batch drifts further from the seed-trained reference than large batch, mirroring the SpecAug signature but driven at the optimizer level rather than the input level.

\begin{figure*}[t]
\centering
\begin{subfigure}[t]{0.28\textwidth}
    \centering
    \includegraphics[width=\linewidth]{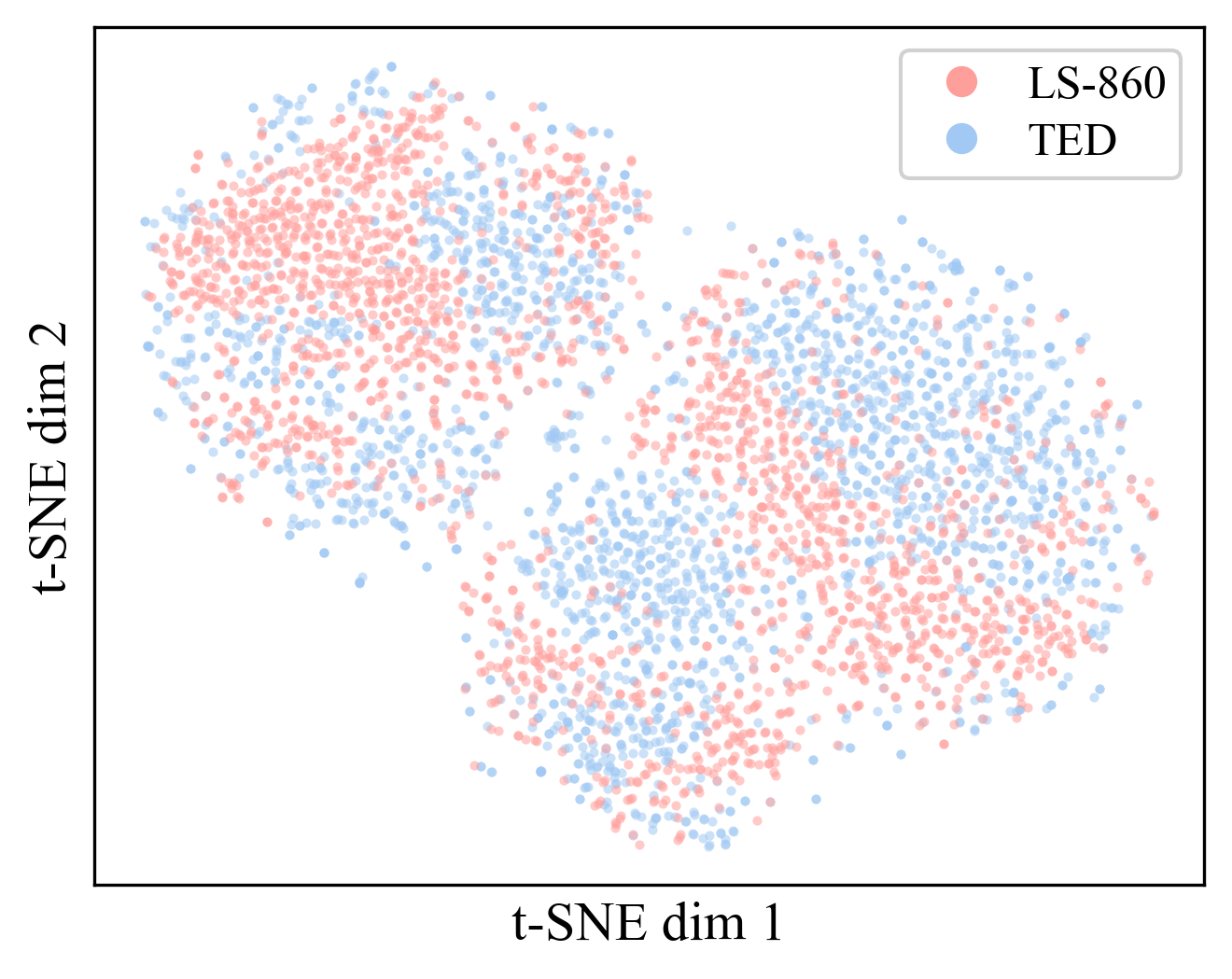}
    \caption{Speaker-embedding t-SNE.}
    \label{fig:dispersion-tsne}
\end{subfigure}
\hfill
\begin{subfigure}[t]{0.7\textwidth}
    \centering
    \includegraphics[width=\linewidth]{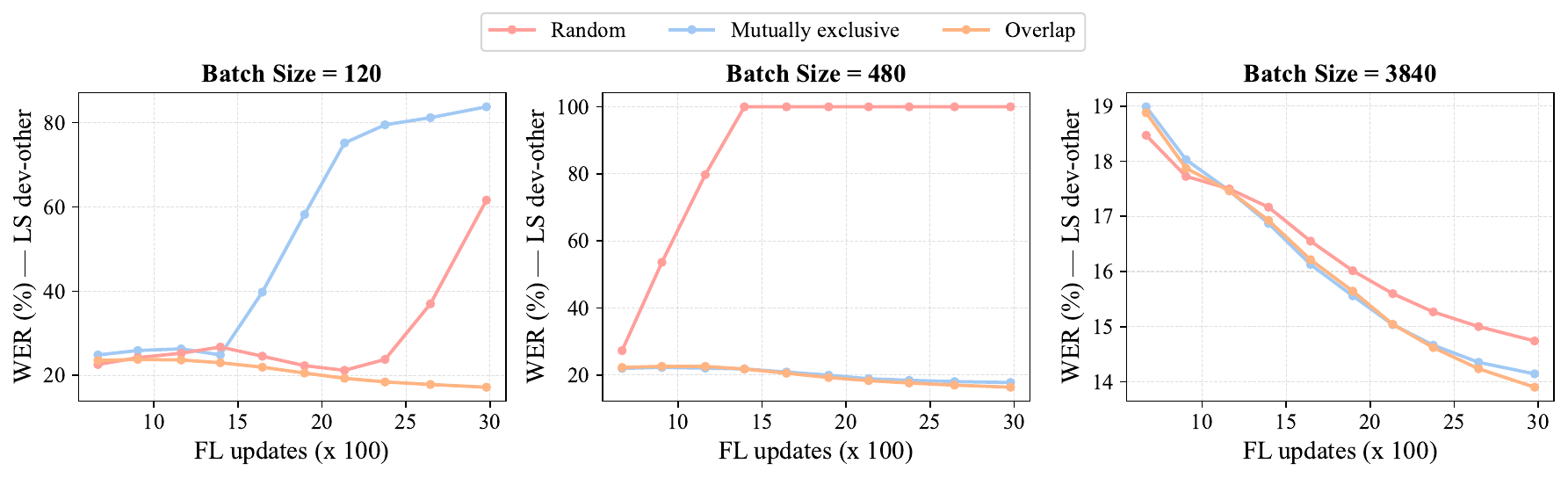}
    \caption{Batch-size ablation.}
    \label{fig:ablation-batch}
\end{subfigure}
\caption{(a)~t-SNE of speaker embeddings for LS860 and Ted samples, visualizing the relative dispersion of the two corpora. 
(b)~Batch-size ablation on Ted $\rightarrow$ LS (LS dev-other). 
Each panel fixes the server batch size ($120$, $480$, $3840$) and overlays WER over FL rounds for three sampling schemes: \emph{random}, \emph{mutually exclusive}, and \emph{overlapped}.
Random diverges or performs worse at every batch size, mutually exclusive diverges only at the smallest batch, and overlapped converges throughout.}
\label{fig:dispersion-batch}
\end{figure*}

\paragraph{Dispersion, not overlap.}
A larger batch could stabilize training for either of two reasons: it lowers the variance of the server gradient by averaging over more samples (a \emph{dispersion} effect), or it is more likely to include samples that resemble the client data, making the supervised update a better proxy for the client distribution (an \emph{overlap} effect).
To separate the two, we vary a batch's dispersion and its overlap with the client distribution independently.
This separation is possible because the seed (Ted) and client (LS860) data form distinct clusters in a t-SNE of their speaker embeddings as shown in \Cref{fig:dispersion-tsne}, which lets us construct batches that either avoid or include client-like samples.
Concretely, we draw server batches in three ways: \emph{random} (high dispersion, some overlap), \emph{mutually exclusive} (low dispersion, no overlap with the client distribution), and \emph{overlapped} (low dispersion, high overlap).
The two hypotheses make opposite predictions. If overlap were the driver, the mutually exclusive scheme---which contains no client-like samples---should diverge while the random and overlapped schemes converge. 
If dispersion were the driver, the two low-dispersion schemes should converge while the high-dispersion random scheme diverges.

\Cref{fig:ablation-batch} compares the three schemes at batch sizes $120$, $480$, and $3840$.
The random scheme diverges or trains noisily at every batch size, the mutually exclusive scheme diverges only at the smallest batch and converges at the moderate and large batches, and the overlapped scheme converges at every batch size.
Because random is the only high-dispersion scheme and the only one to fail across the board, dispersion is the primary driver of instability.
Overlap plays a secondary role: among the two low-dispersion schemes, the overlapped one is more stable, so the absence of client-like samples does contribute to instability---but only at the smallest batch, and it is recovered by a moderately larger batch. 
Overlap therefore matters less than dispersion and can be compensated by batch size.

Both levers: data augmentation (SpecAug) and batch size feed the same server-update noise, so neither alone suffices: the online teacher matches or exceeds global only when both are jointly in their stable range---which we verify across domain-shift pairs next.

\begin{figure*}[t]
\centering
\includegraphics[width=\linewidth]{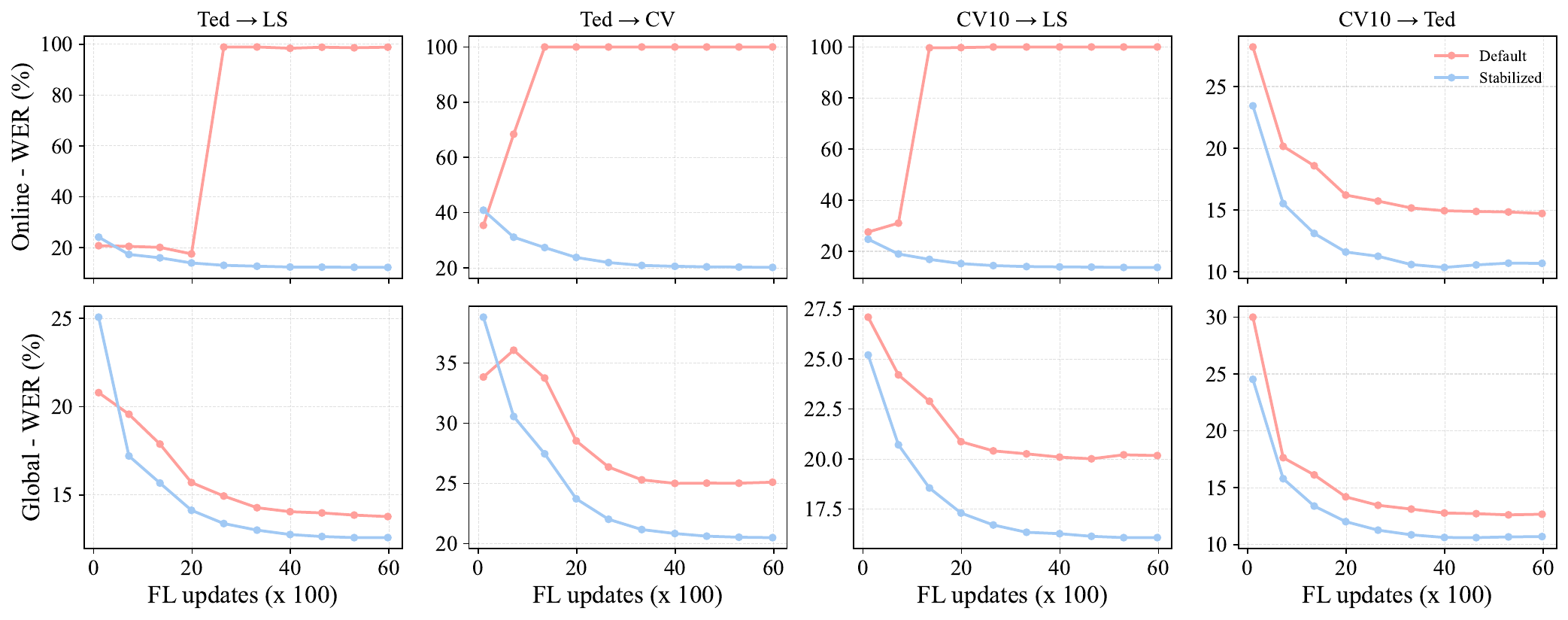}
\caption{Cross-dataset WER comparison across seed $\rightarrow$ client grid. 
Each cell reports \{online, global\} teacher with and without stabilization applied. 
The stabilized transitioning teacher (large-batch, moderate-SpecAug server training) wins on every converging pair.}
\label{fig:cross-dataset}
\end{figure*}

\subsection{Stabilizers Generalize Across Domain-Shift Pairs}
\label{subsec:stabilizers-cross}

We now test whether the two stabilizers identified on Ted $\rightarrow$ LS in \Cref{subsec:why-breaks-server} generalize across multiple cross-domain pairs.
We extend the analysis to four cross-domain pairs: Ted $\rightarrow$ LS, Ted $\rightarrow$ CV, CV10 $\rightarrow$ LS, and CV10 $\rightarrow$ Ted in \Cref{fig:cross-dataset}.
For each pair we cross the pseudo-label source (global or online) with a stabilizer (a $4800$ batch with SpecAug scale $0.5$) or without a stabilizer (a $480$ batch with SpecAug scale $1.0$), yielding four configurations per pair.

Without the stabilizers, the online teacher diverges except for CV10 $\rightarrow$ Ted, while the global teacher stays stable but plateaus at a higher WER than the stabilized online teacher.
With the stabilizers, all online teachers converge. 
Interestingly, the global teachers also benefit from the stabilizers, with significant WER improvements over the non-stabilized global teacher across all pairs.

Together, these results show that the two stabilizers are not specific to the Ted $\rightarrow$ LS direction.
They generalize across all four cross-domain pairs, restoring convergence to the online teacher wherever it previously diverged and lowering WER for the global teacher as well.
Stabilized server training is thus a prerequisite for the online teacher under domain shift, regardless of the particular source and client corpora.
We carry this stabilized configuration into the cross-dataset comparison against existing SSFL methods in \Cref{sec:analysis}.

\begin{takeaways}
\begin{itemize}[leftmargin=*, itemsep=2pt, topsep=2pt]
    \item The online teacher can break, and the failure is \emph{asymmetric}: Ted $\rightarrow$ LS diverges while the reverse LS $\rightarrow$ Ted converges, so it is directional rather than a matter of domain-gap magnitude.
    \item The breakage stems not from seed training but from the server update: the noise it injects---input-level from strong augmentation, optimizer-level from a small batch---pushes the model off-manifold and makes pseudo-labels inconsistent across clients, degrading WER or, in the extreme, diverging.
    \item This instability is stabilized simply, by reducing the data augmentation strength and increasing the batch size, which helps both the online and global teachers and restores convergence even on the pairs where the online teacher previously diverged.
\end{itemize}
\end{takeaways}

\section{Comprehensive Analysis and Practical Guidance}
\label{sec:analysis}

\subsection{Comprehensive Comparison}
\label{subsec:comprehensive-comparison}

\paragraph{Experimental setting.}
\Cref{tab:min-wer-global-vs-online} reports the best WER each method reaches across the seed $\rightarrow$ client grid (11 pairs in total): the supervised upper bound (GT labels), three existing SSFL methods (Static PL, FedNST~\citep{mehmood2022fednst}\footnote{FedNST trains on the full labeled server data each round, making it $1.3$--$2.3\times$ slower than the other methods (\Cref{app:compute}). Within the shared $7$-day wall-clock limit it may not complete the full step budget. \Cref{tab:min-wer-global-vs-online} entries reflect truncated runs.}, and Rao et al.~\citep{rao2023fl_self}) as well as the global and online teachers, each with and without the stabilizer.
To track whether the trained model forgets the seed domain, we also report the source-domain WER in parentheses, which is meaningful only for cross-domain pairs.
For the in-domain pairs, the seed and client are disjoint subsets of the same corpus so the client never sees the labeled seed: LS100 (seed) with LS860 (client) for LS, and CV10 (seed) with CV90 (client) for CV.

To mimic realistic FL training scenarios, we set the cohort size to roughly 1--3\% of the clients in each dataset: 64 for LS and Ted, 1024 for CV, and 256 for Fisher, whose client pools number about 2k, 2k, 35k, and 11k respectively.
We use 20 or 40 local steps, choosing the better per pair on the validation set since the optimum varies across methods, except for LS100 $\rightarrow$ LS860 where we use 160.

The stabilizer is a large-batch server update (gradient accumulation over 10 default batches). For the online teacher we pair it with moderate SpecAugment (scale 0.4 or 0.5), whereas for the global teacher we keep the default mask widths $W_f = 30$ and $W_t = 50$, which we find works best for it.

\paragraph{Results.}
We observe that \citet{rao2023fl_self} is the strongest existing SSFL method, but a large gap to the GT-label upper bound remains, e.g., $7.18$ vs. $14.34$ on CV10 $\rightarrow$ Ted and $10.53$ vs. $42.56$ on CV10 $\rightarrow$ Fisher.
With the stabilizer, both teachers improve over \citet{rao2023fl_self} on most pairs---the global teacher on 8 of 11 and the online teacher on 9 of 11.
Without the stabilizer, the online teacher diverges on 5 of 11 pairs, but with it the online teacher converges on all 11, with the largest gains on high-dispersion (Ted) sources where it recovers from divergence (Ted $\rightarrow$ LS: Div.\ vs. $11.34$, Ted $\rightarrow$ Fisher: Div.\ vs. $18.51$).

Overall, both the stabilized online and global teachers substantially outperform existing SSFL methods, closing much of the gap to the GT-label upper bound.
The stabilized online teacher is the strongest configuration for in-domain pairs, where it adapts quickly to the client distribution and its added noise matters little given the heavy source--client overlap.
Under domain shift, the two teachers reach similar WER, but we favor the global teacher for its robustness: its pseudo-labels stay fixed within an FL round, making it less sensitive to the SpecAugment strength and batch size than the online teacher.

\begin{table}[t!]
\centering
\footnotesize
\begin{tabular}{l ccc cc ccc ccc}
\toprule
\multirow{2}{*}{\textbf{Method}} & \multicolumn{3}{c}{\emph{Client: LS}} & \multicolumn{2}{c}{\emph{Client: Ted}} & \multicolumn{3}{c}{\emph{Client: CV}} & \multicolumn{3}{c}{\emph{Client: Fisher}} \\
\cmidrule(lr){2-4} \cmidrule(lr){5-6} \cmidrule(lr){7-9} \cmidrule(lr){10-12}
            & Ted   & LS100 & CV10  & LS              & CV10  & LS    & Ted   & CV10  & LS    & Ted   & CV10  \\
\midrule
GT Labels & \makecell{6.73\\(7.72)} & 7.39 & \makecell{7.56\\(24.64)} & \makecell{6.46\\(8.00)} & \makecell{7.18\\(29.31)} & \makecell{11.39\\(8.85)} & \makecell{14.60\\(8.68)} & 18.28 & \makecell{10.06\\(8.21)} & \makecell{10.22\\(7.58)} & \makecell{10.53\\(30.57)} \\
\midrule
Static PL & \makecell{14.92\\(7.36)} & 17.71 & \makecell{30.11\\(25.14)} & \makecell{10.22\\(7.63)} & \makecell{33.57\\(28.66)} & \makecell{21.25\\(8.86)} & \makecell{24.90\\(8.64)} & 23.44 & \makecell{31.35\\(7.90)} & \makecell{29.98\\(7.78)} & \makecell{Div.\\(28.68)} \\
FedNST & \makecell{16.59\\(7.34)} & 18.43 & \makecell{30.70\\(19.54)} & \makecell{11.26\\(6.86)} & \makecell{36.68\\(24.75)} & \makecell{22.49\\(6.39)} & \makecell{23.12\\(6.83)} & 23.82 & \makecell{31.58\\(6.58)} & \makecell{28.61\\(6.81)} & \makecell{Div.\\(24.55)} \\
Rao~\etal & \makecell{\textbf{11.12}\\(7.50)} & 15.26 & \makecell{17.91\\(24.97)} & \makecell{8.21\\(7.61)} & \makecell{14.34\\(28.76)} & \makecell{16.12\\(8.28)} & \makecell{19.80\\(8.58)} & 22.32 & \makecell{\textbf{19.82}\\(7.95)} & \makecell{\underline{19.35}\\(7.81)} & \makecell{42.56\\(28.72)} \\
\midrule
Global teacher & \makecell{12.62\\(7.91)} & 12.55 & \makecell{14.84\\(21.59)} & \makecell{8.82\\(7.31)} & \makecell{13.88\\(25.84)} & \makecell{17.55\\(8.60)} & \makecell{20.45\\(7.75)} & 24.78 & \makecell{21.06\\(7.42)} & \makecell{23.10\\(7.59)} & \makecell{33.02\\(27.13)} \\
\quad + stabilizer & \makecell{11.42\\(6.70)} & 13.15 & \makecell{\textbf{13.40}\\(18.43)} & \makecell{\underline{8.08}\\(7.34)} & \makecell{\underline{11.09}\\(22.27)} & \makecell{\underline{15.82}\\(6.88)} & \makecell{\textbf{17.89}\\(6.88)} & \underline{22.21} & \makecell{\underline{20.78}\\(6.98)} & \makecell{25.58\\(6.68)} & \makecell{\underline{29.44}\\(22.64)} \\
Online teacher & Div. & \underline{9.44} & Div. & \makecell{9.12\\(7.54)} & \makecell{15.84\\(27.18)} & \makecell{17.24\\(8.48)} & Div. & 25.65 & \makecell{24.18\\(7.39)} & Div. & Div. \\
\quad + stabilizer & \makecell{\underline{11.34}\\(6.75)} & \textbf{9.25} & \makecell{\underline{13.50}\\(18.53)} & \makecell{\textbf{7.79}\\(7.40)} & \makecell{\textbf{11.07}\\(22.53)} & \makecell{\textbf{15.58}\\(6.79)} & \makecell{\underline{18.85}\\(6.92)} & \textbf{21.84} & \makecell{21.45\\(7.19)} & \makecell{\textbf{18.51}\\(6.90)} & \makecell{\textbf{27.43}\\(26.05)} \\
\bottomrule
\end{tabular}
\caption{WER (\%) on the test sets across eleven (seed, client) pairs, grouped by client dataset (top header), with each column a seed dataset. Each cell shows the target-domain WER on top and, where available, the source-domain WER in parentheses below. \emph{Div.}\ denotes divergence.}
\label{tab:min-wer-global-vs-online}
\end{table}

\subsection{Transitioning Teacher}
\label{subsec:transitioning-teacher}

\begin{wrapfigure}{r}{0.38\textwidth}
    \vspace{-0.52cm}
    \centering
    \includegraphics[width=0.32\textwidth]{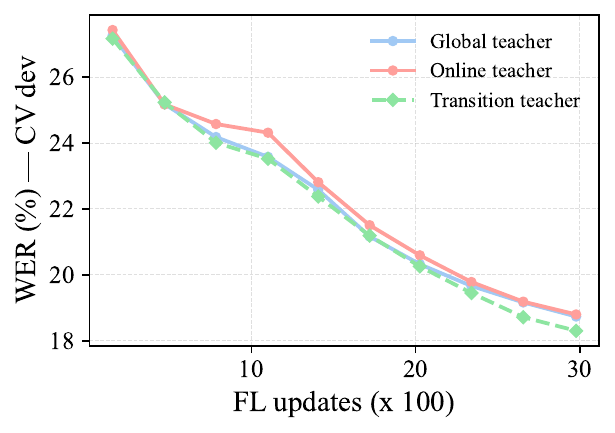}
    \vspace{-0.39cm}
    \caption{
        Teachers with the stabilizer
        }
    \label{fig:transitioning-teacher}
    \vspace{-0.3cm}
\end{wrapfigure}

We further validate the effectiveness of the transitioning teacher, which switches from the global to the online teacher at a fixed round $r$, on the LS960 $\rightarrow$ CV pair with the stabilizers applied.
Unlike \Cref{fig:transition-curves} where the stabilizers are not applied, here we apply them to all teachers.

As \Cref{fig:transitioning-teacher} shows, the stabilizers shrink the gap between the global and online teachers considerably (without the stabilizers, global: $20.94$ vs. online: $21.55$, and with the stabilizers, global: $18.65$ vs. online: $18.73$ WER).
The transitioning teacher (switching at $r=2000$) still outperforms both, though by a smaller margin than in \Cref{fig:transition-curves} because the stabilizers already improve the global and online teachers significantly (transitioning: $19.7$ without the stabilizers vs. $18.25$ with them).
In summary, the transitioning teacher yields a small but consistent improvement by recovering the online teacher's early-round disadvantage, capturing the global teacher's early stability and the online teacher's late adaptivity.
Since this margin is modest once the stabilizer is applied, we present it as an analysis of that early-round effect rather than as a general recommendation.

\begin{figure*}[t]
\centering
\begin{subfigure}[t]{0.48\textwidth}
    \centering
    \includegraphics[width=\linewidth]{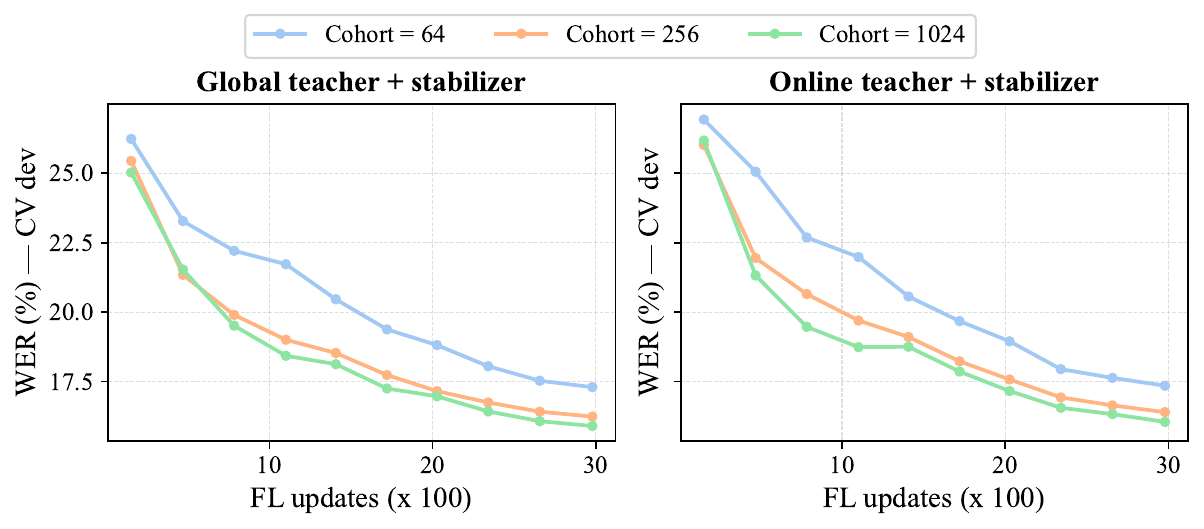}
    \caption{Cohort size.}
    \label{fig:ablation-cohort}
\end{subfigure}
\hfill
\begin{subfigure}[t]{0.48\textwidth}
    \centering
    \includegraphics[width=\linewidth]{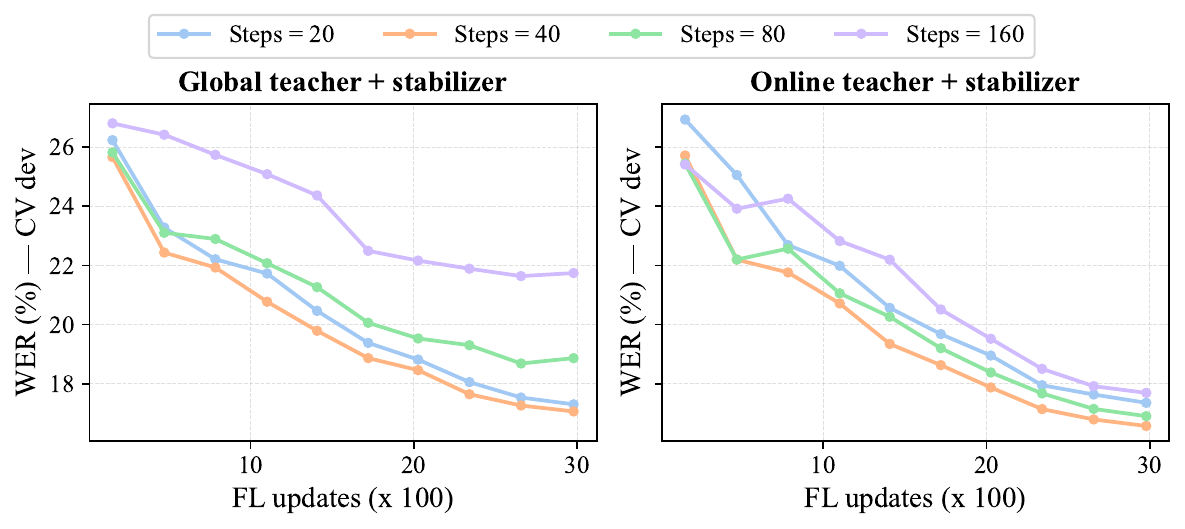}
    \caption{Local steps.}
    \label{fig:ablation-localsteps}
\end{subfigure}
\caption{Ablation studies on LS $\rightarrow$ CV for the global and online teachers, both with stabilizers. (a)~Effect of cohort size $\{64, 256, 1024\}$ at 20 local steps. (b)~Effect of the number of local steps $\{20, 40, 80, 160\}$ at cohort size 64.}
\label{fig:ablation}
\end{figure*}
\subsection{Ablation Studies}
\label{subsec:ablations}

\paragraph{Cohort size.}
Cohort size---the number of clients sampled per round---is a primary knob of FL, and its effect on SSFL may not match its effect on supervised FL, since there is additional noise coming from pseudo-labeling and server updates.
We ablate it for the global and online teachers, both with stabilizers, over cohort sizes $\{64, 256, 1024\}$ on LS $\rightarrow$ CV at 20 local steps in \Cref{fig:ablation-cohort}.
We observe that as cohort size increases, the performance increases for both global and online teachers.
There is no significant difference in trend between global and online teachers.
It is generally advisable to maximize cohort size if possible.

\paragraph{Local training steps.}
Another important knob for FL training is the number of local steps each client takes per round. 
It governs the communication--computation trade-off, but in SSFL, it also controls how much the online teacher adapts within a round, and hence the quality of the pseudo-labels it produces. 

We ablate it for the global and online teachers, both with stabilizers, over $\{20, 40, 80, 160\}$ local steps on LS $\rightarrow$ CV at cohort size 64 in \Cref{fig:ablation-localsteps}.
For both global and online teachers, 40 local steps are the best choice. If the number of local steps is too large, the performance degrades for both teachers, for different reasons.
As the number of local steps increases, pseudo-labels produced by the global teacher become stale, which leads the local model to update in a suboptimal direction.
On the other hand, the online teacher, although it does not produce stale pseudo-labels, overfits to the client data too much, which induces severe drift between client models and destabilizes aggregation.

\paragraph{Data augmentation and dropout.}
We generate pseudo-labels without SpecAugment or dropout to give the teacher a clean signal.
During local training, however, we apply both SpecAugment and dropout to prevent overfitting to the pseudo-labels, following common practice.
It is unclear how SpecAugment and dropout interact with the server update probability $p$ and affect the online teacher, so we ablate them in \Cref{tab:probsup-specaug} on both in-domain (LS100 $\rightarrow$ LS860) and cross-domain (LS $\rightarrow$ CV) settings at cohort size 64 with 20 local steps.

As in \Cref{fig:pl-source-basics-failure}, at $p=0.0$ the online teacher diverges in both settings.
At $p=0.2$ it tends to converge in both, though it still diverges in some cases, and when it does converge its WER is lower than at $p=0.5$.
This is because a lower $p$ makes training rely more on the client data than the server data, which improves WER.
Lowering $p$ too far, however, can cause divergence due to client drift.
Dropout as regularization partly offsets this risk: although $p=0.2$ generally diverges without dropout in-domain, it converges with dropout.

We observe a similar trend for SpecAugment strength in-domain: a higher scale improves WER but raises the chance of divergence, which dropout again mitigates.
For cross-domain, the effect of SpecAugment is less pronounced as the optimal scale is lower than in-domain.
When SpecAugment is too strong, the online teacher diverges in both settings whereas when it is too weak, the WER can be higher.
Therefore, we recommend that practitioners jointly tune SpecAugment strength and the server update probability $p$ along with dropout as regularization to minimize WER in their own applications.

\begin{table}[t]
\centering
\setlength{\tabcolsep}{3pt}
\begin{subtable}{0.49\textwidth}
\centering
\begin{minipage}[t]{0.48\linewidth}
\centering
\textbf{\small without dropout}\\[3pt]
\begin{tabular}{lccc}
\toprule
 & \multicolumn{3}{c}{$p$} \\
\cmidrule(lr){2-4}
Scale & $0.5$ & $0.2$ & $0.0$ \\
\midrule
$1.0$ & Div. & Div. & Div. \\
$0.3$ & 14.1 & Div. & Div. \\
$0.0$ & 15.3 & 11.7 & Div. \\
\bottomrule
\end{tabular}
\end{minipage}
\hfill
\begin{minipage}[t]{0.48\linewidth}
\centering
\textbf{\small with dropout}\\[3pt]
\begin{tabular}{lccc}
\toprule
 & \multicolumn{3}{c}{$p$} \\
\cmidrule(lr){2-4}
Scale & $0.5$ & $0.2$ & $0.0$ \\
\midrule
$1.0$ & 11.0 & 10.3 & Div. \\
$0.3$ & 12.3 & 10.8 & Div.\\
$0.0$ & 12.9 & 10.4 & Div. \\
\bottomrule
\end{tabular}
\end{minipage}
\caption{In-domain: LS100 $\rightarrow$ LS860.}
\label{tab:grid-indomain}
\end{subtable}
\hfill
\begin{subtable}{0.49\textwidth}
\centering
\begin{minipage}[t]{0.48\linewidth}
\centering
\textbf{\small without dropout}\\[3pt]
\begin{tabular}{lccc}
\toprule
 & \multicolumn{3}{c}{$p$} \\
\cmidrule(lr){2-4}
Scale & $0.5$ & $0.2$ & $0.0$ \\
\midrule
$1.0$ & 29.8 & Div. & Div. \\
$0.3$ & 24.6 & 21.7 & Div. \\
$0.0$ & 24.9 & 21.7 & Div. \\
\bottomrule
\end{tabular}
\end{minipage}
\hfill
\begin{minipage}[t]{0.48\linewidth}
\centering
\textbf{\small with dropout}\\[3pt]
\begin{tabular}{lccc}
\toprule
 & \multicolumn{3}{c}{$p$} \\
\cmidrule(lr){2-4}
Scale & $0.5$ & $0.2$ & $0.0$ \\
\midrule
$1.0$ & 26.2 & Div. & Div. \\
$0.3$ & 24.5 & 21.8 & Div. \\
$0.0$ & 24.7 & 21.6 & Div. \\
\bottomrule
\end{tabular}
\end{minipage}
\caption{Cross-domain: LS $\rightarrow$ CV.}
\label{tab:grid-crossdomain}
\end{subtable}
\caption{Scale for SpecAugment strength ($1.0/0.3/0.0$, rows) $\times$ server update probability $p$ ($0.5/0.2/0.0$, columns) grids of final WER (\%), for in-domain and cross-domain settings, without and with dropout. \emph{Div.}\ denotes divergence.}
\label{tab:probsup-specaug}
\vspace{-5mm}
\end{table}

\subsection{When to Use What: Practical Guidelines}
\label{subsec:guidelines}

\vOneEdits{These guidelines are derived from the public-benchmark experiments of \Cref{sec:setup}; \Cref{sec:discussion} discusses their scope and limitations.} Our findings translate into a simple recipe that requires no per-pair hyperparameter tuning, driven by two properties a practitioner can assess up front: whether the setting is in-domain or cross-domain, and the dispersion and audio duration of the server (seed) data.

\paragraph{Pseudo-label source.}
The choice of teacher follows from the overlap between the server and client data, which can usually be judged from domain knowledge.\footnote{When domain knowledge is insufficient, this overlap can be estimated directly from data: the embedding-based similarity measure of \Cref{app:domain-shift} separates in-domain from cross-domain pairs and serves as a quantitative proxy for it.}
When the two are highly overlapped (in-domain), we recommend the online teacher, whose within-round adaptation to the client distribution improves the pseudo-labels while its added noise matters little.
Under domain shift, the online and global teachers reach similar WER on most pairs, but we recommend the global teacher as the safer default. Because its pseudo-labels stay fixed within a round, it is more robust to the choice of SpecAugment strength and batch size, whereas the online teacher requires careful tuning of these settings to avoid divergence.

\paragraph{Server update settings.}
We recommend keeping the server update as a simple supervised batch step on the seed data, applied on its own between FL rounds, rather than folding it into the client aggregation as FedNST~\citep{mehmood2022fednst} and Rao~\etal~\citep{rao2023fl_self} do (\Cref{app:baselines}).
This plain update is simpler to implement and tune.
It is also no less effective in our comparison, where neither folded-update baseline reaches a lower WER (\Cref{tab:min-wer-global-vs-online}).

Two further settings then control its stability and should be matched to the source data.
First, the server batch size should scale with the dispersion of the server data: a higher-dispersion source needs a larger batch to keep the variance of the server gradient low enough to avoid divergence (\Cref{subsec:why-breaks-server}).
This adjustment helps both teachers.
Second, the SpecAugment strength interacts with the teacher choice.
For the online teacher, it should be calibrated to the audio duration of the source, since longer clips tolerate stronger masking whereas shorter clips require weaker masking to keep the effective mask coverage below the divergence threshold.
The global teacher, by contrast, is far less sensitive to SpecAugment strength and generally benefits from strong masking.

\begin{takeaways}
\begin{itemize}[leftmargin=*, itemsep=2pt, topsep=2pt]
    \item Across the 11 seed--client pairs, the global and online teachers \emph{with the stabilizer} substantially outperform existing SSFL methods (Static PL, FedNST, and Rao et al.), closing much of the gap to fully-supervised FL.
    \item No single teacher dominates: the online teacher wins in-domain and on most cross-domain pairs, but under domain shift the two are close, so the more robust global teacher---less sensitive to the data augmentation and batch-size settings---is the safer choice.
    \item These findings yield a simple recipe that needs no per-pair tuning: (1)~use the online teacher in-domain and the global teacher under domain shift, (2)~keep a plain supervised batch update on the server, (3)~scale the server batch size with the dispersion of the server data, and (4)~calibrate the SpecAugment strength (to audio duration for the online teacher).
\end{itemize}
\end{takeaways}

\section{Discussion}
\label{sec:discussion}

\subsection{Limitations}
This work demonstrates the advantage of different pseudo-label teachers and the impact of server-side stabilizers. However, these findings may not transfer to settings that differ from our experimental setup. We outline the main limitations in scope below so that practitioners can judge how well our conclusions apply to their own settings.

\paragraph{Language and dataset scope.}
All four corpora we study (LS, Ted, CV, and Fisher) are English. 
Languages that are tonal, low-resource, or written in non-Latin scripts may exhibit different SpecAugment-coverage thresholds and a different online-vs.-global trade-off.

\paragraph{Architecture and training scope.}
Our experiments use a single architecture, a Transformer encoder with a CTC head. Other families---RNN-T~\citep{graves2012transducer}, attention encoder--decoder~\citep{chan2016las}, and Conformer~\citep{gulati2020conformer}---may express the two server-training failure modes (divergence under strong SpecAugment versus higher-WER instability under small batches) differently. 
We also decode greedily without a language model, whereas beam search with an external LM is standard in deployment, so whether the online-teacher advantage compounds or diminishes under LM-fused decoding is untested. 
Finally, we fix the optimizer pair to SGD on clients and LAMB~\citep{you2020lamb} on the server. 
Adam-class server optimizers such as FedAdam~\citep{reddi2021fedopt} may interact differently with the SpecAugment strength.

\paragraph{Federated learning setup.}
\vOneEdits{All experiments use public benchmark corpora (LibriSpeech, TED-LIUM, Common Voice, Fisher); no real user data, production traffic, or production-system telemetry informed any part of this study, and the reported system parameters (cohort sizes, compute budgets, communication-round counts, optimizer settings) are not intended to characterize, and have not been validated against, any production configuration.} We simulate clients by sampling cohorts from static datasets, so the non-stationary data, mid-round dropout, network failures, and continuous churn of real deployments are not modeled.
In addition, the recommended stabilizer configuration with a large batch, e.g., a batch of 480 accumulated over 10 steps, requires substantial server-side compute.
Although server compute is often not a bottleneck in practice, practitioners with constrained server budgets may not be able to apply the recipe directly.

\paragraph{Baseline comparison scope.}
Our comparison against FedNST and Rao~\etal\ (\Cref{tab:min-wer-global-vs-online}) is end-to-end rather than a controlled ablation of the server-update mechanism, since these baselines also differ from our configuration in their pseudo-label teacher.
We therefore recommend keeping the server update as a simple supervised batch step applied on its own between FL rounds (\Cref{subsec:guidelines}) primarily for its simplicity, and leave a controlled comparison to future work.

\subsection{Future Work}

\paragraph{Privacy guarantee.}
Our recipe leaves the FL privacy guarantee intact: the online teacher generates pseudo-labels locally with no extra communication, and the larger server batches use only the server's own labeled data. We have not, however, tested it under differential privacy. Since DP-FL adds per-layer gradient clipping and noise while our stabilizer relies on large, low-variance server updates, the two may work against each other---DP noise could undo the variance reduction the stabilizer provides, so batch size and SpecAugment strength may need re-tuning under a DP budget. Quantifying this interaction, and establishing formal guarantees under a specified threat model, is a natural next step.

\paragraph{Better decoding and data filtering.}
We decode greedily and apply no pseudo-label filtering---two axes orthogonal to the teacher and stabilizer choices studied here that could be layered on top. Beam search with an external language model is standard in deployment, and confidence-based filtering of pseudo-labels (held off throughout, \Cref{subsec:constraints}) is a common SSL ingredient. Whether the online-teacher advantage compounds with LM-fused decoding, and whether confidence thresholds compose with the transitioning teacher and the stabilizers, are open questions worth pursuing.

\section{Conclusion}
\label{sec:conclusion}

We studied semi-supervised federated learning (SSFL) for automatic speech recognition (ASR), a setting in which pseudo-label errors compound across the output sequence and across training rounds into divergence and leave a large gap to fully-supervised FL. 
We showed that closing this gap turned on two coupled design axes---the teacher that generates the pseudo-labels and the anchor that stabilizes training through server-side updates on labeled data. 

On the \textit{teacher axis}, the best choice was dynamic rather than fixed: a per-client online teacher matched or beat the broadcast global teacher once stabilized, and a transitioning teacher (global $\rightarrow$ online) outperformed either as the seed model grew stronger. 
On the \textit{anchor axis}, interleaving server training on the labeled seed between FL rounds was a prerequisite for stability, and this anchor---more than the seed model itself---governed convergence. 
The two axes proved inseparable: the aggressive teacher choices paid off only once the anchor stabilized training, which we found was highly sensitive to SpecAugment strength and batch size, the two levers that control server-update gradient noise and hence pseudo-label consistency. 
How much stabilization was needed proved domain-dependent, governed by the dispersion of the seed data and its overlap with the client data.

Together, these findings yielded practical guidelines that stabilized an otherwise brittle procedure and closed much of the gap to fully-supervised FL: applied correctly, the recipe improved WER over the strongest prior SSFL method by $20.8\%$ in-domain and $10.0\%$ cross-domain on average (on 9 of 11 pairs), and by roughly $48\%$ over a naive static pseudo-labeling baseline in-domain ($17.7 \rightarrow 9.3$ WER).

\section*{Acknowledgments}
We thank David Grangier, Skyler Seto, Amar Subramanya, and Russ Web for essential feedback on the paper and Apple infrastructure team for assistance with developing scalable, fault tolerant code.

%% >>>>> begin inlined main-apple.bbl

%% <<<<< end inlined main-apple.bbl
\clearpage

\beginappendix
\crefalias{section}{appendix}

\section{Experimental Details}
\label{app:setup}

This appendix lists the full dataset, model, and federated-learning configuration summarized in \Cref{sec:setup}.
Defaults apply throughout the paper unless a specific section notes that a value is varied. \vOneEdits{These values reflect choices made for our public-dataset simulations and should not be read as characterizing any production system's configuration.}

\subsection{Datasets and Splits}
\label{app:datasets}

\Cref{tab:datasets} summarizes the four English ASR corpora used in our experiments, chosen to span a broad range of speaking styles and recording conditions---from clean read audiobooks to conversational telephone speech---so that our conclusions do not hinge on a single acoustic domain. This appendix describes each corpus, how it is partitioned into server (seed) and client data, the seed and client subsets and domain-shift pairs we evaluate, and the train/dev/test protocol.

\begin{table}[h]
\centering
\small
\caption{Per-corpus statistics. Hours are computed from the per-utterance durations of the audio used in our pipeline. Speaker counts are approximate client-pool sizes, since each speaker is treated as one federated client. Min, median, and max utterance durations are in seconds.}
\label{tab:datasets}
\begin{tabular}{llrrrrr}
\toprule
\textbf{Corpus} & \textbf{Speaking style} & \textbf{Hours} & \textbf{Speakers} & \textbf{Min\ (s)} & \textbf{Median\ (s)} & \textbf{Max\ (s)} \\
\midrule
LibriSpeech (LS)  & Read audiobooks        & 960  & $\sim$2k  & 0.8 & 13.8 & 29.7 \\
TED-LIUM (Ted)    & TED talks              & 452  & $\sim$2k  & 0.1 & 5.9  & 30.3 \\
Common Voice (CV) & Crowd-sourced read     & 1593  & $\sim$35k & 1.4 & 5.6  & 13.0 \\
Fisher            & Telephone conversation & 1928 & $\sim$11k & 0.3 & 2.4  & 203.0 \\
\bottomrule
\end{tabular}
\end{table}

\paragraph{Corpora.}
\textbf{LibriSpeech (LS)} consists of read English audiobooks sampled at 16\,kHz, yielding long, fluent utterances with few disfluencies.
\textbf{TED-LIUM (Ted)} contains prepared but spontaneously delivered TED-talk speech, with hesitations, restarts, and a single speaker per talk.
\textbf{Common Voice (CV)} is a large crowd-sourced corpus of short read sentences recorded by many volunteers on heterogeneous consumer devices, giving wide accent and channel diversity.
\textbf{Fisher} comprises two-party conversational English telephone speech sampled at 8\,kHz, with short, disfluent, and often overlapping turns.
Together they cover read, spontaneous, crowd-sourced, and telephone speech, which lets us study domain shift across genuinely different acoustic and linguistic conditions.

\paragraph{Server and client partitioning.}
Each corpus is used either as a labeled server (seed) corpus or as an unlabeled client corpus, depending on the pair.
For client corpora, we partition the data by speaker and treat each speaker as a separate \vOneEdits{simulated} federated client.
For server corpora, we do not partition the data\footnote{The exception is our reproduction of \citet{rao2023fl_self}, for which we need rehearsal data partitioned by speaker ID.} and instead sample training batches uniformly at random.
In both roles, utterances longer than 30\,s are filtered out before training (\texttt{max\_audio\_len\_s}=30.0).

\paragraph{Seed and client subsets.}
We denote by LS$x$ an $x$-hour labeled subset of LS used as the seed corpus, with the disjoint remainder serving as the unlabeled client corpus.
For example, LS100 $\rightarrow$ LS860 pairs a 100\,h seed with the remaining 860\,h as clients, and the seed-strength sweep in \Cref{subsec:seed-strength} uses LS100, LS360, LS600, and LS960 seeds.
Similarly, CV10 and CV90 denote a 10\%/90\% split of CV into seed and client partitions.

\paragraph{Domain-shift pairs.}
We evaluate the eleven server $\rightarrow$ client pairs listed in \Cref{tab:min-wer-global-vs-online}.
They range from \emph{in-domain} pairs, where the seed and client are disjoint splits of the same corpus (LS100 $\rightarrow$ LS860 and CV10 $\rightarrow$ CV90), through \emph{moderate} shifts between read and spontaneous speech (LS $\leftrightarrow$ Ted), to \emph{large} shifts onto crowd-sourced or telephone speech (e.g., LS $\rightarrow$ CV at several seed strengths, CV10 $\rightarrow$ Ted, and pairs involving Fisher).
Where possible we evaluate both directions of a shift (e.g., LS $\rightarrow$ Ted and Ted $\rightarrow$ LS), which lets us test whether the domain gap acts symmetrically.

\paragraph{Train/dev/test splits.}
We use the standard benchmark splits for each corpus.
For LS we use dev-clean and dev-other (and test-clean and test-other for \Cref{tab:min-wer-global-vs-online}), for CV the English dev and test sets, and for Ted its dev and test sets.
As noted in \Cref{subsec:datasets}, all analysis and hyperparameter-tuning results are reported on the dev sets, and only the final comprehensive comparison in \Cref{tab:min-wer-global-vs-online} uses the held-out test sets, which keeps the test data untouched during model development.

\subsection{Model and Training Hyperparameters}
\label{app:model}

\subsubsection*{Model Parameters}
\begin{description}[leftmargin=0pt, style=unboxed, font=\normalfont\bfseries]
    \item[Architecture.] The acoustic model is a Transformer encoder~\citep{vaswani2017transformer} with 36 blocks, a hidden/embedding
  dimension of 768, 4 attention heads, and a feed-forward (MLP) dimension of 3072, preceded by a
  convolutional front-end (kernel 7, stride 3) that subsamples the input by a factor of 3. The model has
  approximately 255M parameters.
    \item[Audio features.] We extract 80-dimensional log-mel filterbank features using a 25\,ms window and
  a 10\,ms stride. Utterances whose duration falls outside $[0\,\text{s}, 30\,\text{s}]$ are discarded
  (\texttt{max\_audio\_len\_s}=30.0), and the maximum target length is capped at 400 characters.
    \item[Loss.] The model is trained with the connectionist temporal classification (CTC) loss~\citep{graves2006ctc}.
    \item[Tokenizer.] We use a character-level tokenizer with an output vocabulary of 28 characters plus
  the CTC blank symbol.
    \item[Decoding.] Decoding is greedy (argmax over the CTC posteriors) and uses no external language
  model.
\end{description}

\subsubsection*{Training Parameters}
\begin{description}[leftmargin=0pt, style=unboxed, font=\normalfont\bfseries]
    \item[Augmentation.] We apply SpecAugment~\citep{park2019specaugment} with 2 frequency masks of maximum
  width $W_f = 30$ and 10 time masks of maximum width $W_t = 50$, with the time-mask width capped at a
  ratio of $0.1$ of the utterance length and no mask averaging. Augmentation is enabled from the first
  training step (\texttt{start\_saug}=0).
    \item[Dropout.] We use dropout rate of 0.3 to train the seed model in the separate central run. During FL training, Dropout is fixed to 0.1 throughout, with \texttt{dropout} = \texttt{layer\_dropout} =
  0.1 to increase the capacity of the model following \citet{likhomanenko2021slimipl}. 
    \item[Client optimizer.] Clients optimize with SGD at a learning rate of 0.2 for Ted sources and
  0.4 for LS sources, without any learning-rate schedule.
    \item[Server optimizer.] %
    The server optimizes with LAMB~\citep{you2020lamb} using an exponential-decay schedule with decay rate 0.6. The server learning rate is 0.004 for most source$\to$target pairs, with the exceptions of Ted$\to$LS and LS-100$\to$LS-860 (0.003) and CV-10$\to$CV-90 (0.001). 
    The schedule is set relative to the total number of FL steps $T$: for most pairs ($T = 3000$), decay begins at $T/3$ and the learning rate is scaled by $0.6$ for every subsequent $T/6$ steps. 
    The two exceptions use $T = 2000$: LS-100$\to$LS-860 begins decay at $T/2$ and is scaled by $0.6$ every $T/8$ steps, whereas CV-10$\to$CV-90 begins decay only at $5T/8$.
    \item[Seed training.] The model is seeded by pretraining on the labeled server corpus in a separate
  central run. This checkpoint is loaded to initialize both the student and the EMA teacher.
  \end{description}

\subsection{Semi-Supervised Federated Learning Hyperparameters}
\label{app:fl}

\subsubsection*{Federated Learning Parameters}
\begin{description}[leftmargin=0pt, style=unboxed, font=\normalfont\bfseries]
    \item[Aggregation.] Throughout all experiments, client updates are aggregated using FedAvg~\citep{mcmahan2017fl} with equal weights to all clients that participate.
    \item[FL steps.] We run $T = 3000$ FL steps for most pairs, and $T = 2000$ for the in-domain LS100 $\rightarrow$ LS860 and CV10 $\rightarrow$ CV90 pairs, where one FL step corresponds to a single round of client aggregation and, with probability $p$, a server update. The server learning-rate schedule scales with $T$ as described above.
    \item[Cohort size.] For the comprehensive comparison in \Cref{tab:min-wer-global-vs-online}, we fix the cohort size per client corpus to roughly 1--3\% of its client pool: 64 for LS and Ted, 1024 for CV, and 256 for Fisher. We separately ablate the cohort size over $\{64, 256, 1024\}$ in \Cref{subsec:ablations}.
    \item[Local steps.] Each client performs $K$ local steps per round. By default we use $K = 20$ or $40$, whichever performs better on the validation set, except when the client corpus is LS where we use $K = 160$. We ablate $K$ over $\{20, 40, 80, 160\}$.
    \item[Server training.] In each round, the server performs a supervised pass over the seed corpus with
  probability $p = 0.2$. The default server batch size is 120\,s of audio for LS sources and
  480\,s for Ted sources, and high-dispersion sources use $10\times$ gradient accumulation
  (\texttt{server\_grad\_accum\_steps}=10).
\end{description}

\subsubsection*{Pseudo-Labeling Parameters}
\begin{description}[leftmargin=0pt, style=unboxed, font=\normalfont\bfseries]
    \item[EMA teacher.] The EMA teacher uses a decay rate of $\lambda = 0.99$ (the global-EMA default,
  which the student configs do not override).
    \item[PL source.] Pseudo-labels are produced by a global or online teacher
  (\texttt{cache\_update\_method}=pre with \texttt{use\_ema}). When a transitioning teacher is used, the
  transition round is $r \approx 2000$.
  \end{description}

Unless otherwise stated, all experiments use the defaults above. Each section in the main text explicitly notes which hyperparameters are varied.

\section{Baseline Method Details}
\label{app:baselines}

This appendix expands on how the existing SSFL for ASR baselines instantiate the two design axes of \Cref{subsec:problem-setting}: the pseudo-label source and the server training strategy.
We write $\serverD$ for the labeled server corpus, $\clientD_i$ for the unlabeled data of client $i$, and $\theta_0$ for the seed model trained on $\serverD$.

\subsection{FedNST}
\label{app:fednst}

FedNST~\citep{mehmood2022fednst} extends noisy student training~\citep{park2020nst} to the federated setting.
It first trains a seed model $\theta_0$ on the labeled server data $\serverD$, then runs $T$ federated rounds.

\paragraph{Pseudo-label generation.}
Before federated training begins, every client generates pseudo-labels once from the seed model $\theta_0$, decoding its unlabeled audio $\clientD_i$ with an external language model and beam search\footnote{As mentioned in \Cref{sec:ssfl-asr}, we do not apply any language model fusion or rescoring.}, and caches the resulting transcripts on-device for reuse in every subsequent round.
The teacher is therefore frozen at the seed, and no confidence filtering or class balancing is applied by default.
FedNST also considers regenerating pseudo-labels each round from the latest global model, but this raises the wall-clock cost of a run by roughly $10\times$ for a negligible change in WER, so the one-time variant is used throughout.

\paragraph{Server update.}
FedNST merges the federated aggregation and the server-side supervised update into a single global step.
In round $t$, each sampled client $i$ trains for several local epochs on its cached pseudo-labels and returns a pseudo-gradient $g_t^{i} = \theta_t - \theta_t^{i}$, where $\theta_t^{i}$ are its locally updated parameters.
The server aggregates these FedAvg-style, weighted by the per-client sample count $n_i$,
\begin{equation}
    g_C = \sum_{i} \frac{n_i}{n}\,g_t^{i}, \qquad n = \sum_i n_i,
\end{equation}
and in parallel performs a supervised pass on the labeled server data $\serverD$ to obtain a server pseudo-gradient $g_S$.
The two are combined by a weighted average and applied as one update,
\begin{equation}
    g = \alpha\,g_S + (1-\alpha)\,g_C, \qquad \theta_{t+1} = \textsc{ServerOpt}(\theta_t, g),
\end{equation}
with mixing weight $\alpha = 0.5$ in their experiments (we also use this value throughout the experiments).
Because the labeled pass contributes to every round, the seed signal re-anchors training continuously rather than periodically.

\subsection{Rao et al.}
\label{app:rao}

\citet{rao2023fl_self} study federated continual learning for an RNN-T ASR model, where a paired teacher labels on-device audio and the model is updated without ground-truth transcripts.

\paragraph{Pseudo-label generation.}
Pseudo-labels are produced each round by a paired teacher $\teacher_t$ that is an exponential moving average (EMA) of the global student, refreshed every $u$ rounds,
\begin{equation}
    \teacher_t = \lambda\,\teacher_{t-1} + (1-\lambda)\,\student_t \quad \text{when } t \equiv 0 \pmod{u},
\end{equation}
and held fixed otherwise.
We set $u = 1$ in our experiments, so the teacher is refreshed every round.
Unlike FedNST's frozen seed teacher, this teacher tracks the evolving global model.
Each sampled device transcribes its unlabeled audio $\clientD_i$ with $\teacher_t$, filters out utterances of very low or very high confidence, and trains on the retained samples with audio augmentation.
Optionally, weak-supervision signals such as alternate-system NLU semantics or user feedback scores are folded in through a policy-gradient loss, but these are orthogonal to the pseudo-label source itself.\footnote{In our comparison we apply neither the confidence filtering nor the weak supervision, isolating \citet{rao2023fl_self}'s pseudo-label source and server training strategy.}

\paragraph{Server update.}
The labeled server data enters training as \emph{rehearsal} on a set of cloud ``pseudo-devices'' $\clients_C$.
These pseudo-devices draw ground-truth-labeled data $\serverD$ and compute local updates exactly like real clients.
In round $t$, every participant $k$ in the union of the sampled real devices $\clients_t$ and the cloud pseudo-devices $\clients_C$ returns a pseudo-gradient $g_t^{k} = \student_t - w_t^{k}$, where $w_t^{k}$ are its locally updated parameters, and the server averages them in a single aggregation step,
\begin{equation}
    \student_{t+1} = \textsc{ServerOpt}\!\left(\student_t,\; \frac{1}{|\clients_t \cup \clients_C|}\sum_{k \in \clients_t \cup \clients_C} g_t^{k}\right).
\end{equation}
The supervised signal is therefore not a separate server pass but an additional group of clients folded into the same FedAvg aggregation, acting as a regularizer that mitigates catastrophic forgetting on the seed distribution.
In their experiments, $|\clients_C| = 40$ cloud pseudo-devices are used alongside $|\clients_t| = 400$ sampled real devices per round, a rehearsal cohort of $10\%$ of the sampled clients.
Following this setup, we set the rehearsal cohort to $10\%$ of the FL cohort each round.

\begin{table}[h!]
\centering
\small
\caption{Per-round, per-GPU server-update compute relative to a global/online (ours) step, for LS100$\to$LS860 ($C{=}32, K{=}160, G{=}16$) and LS$\to$CV ($C{=}1024, K{=}20, G{=}64$).
$n_S = |\serverD|/b_S$ is the batch count of one full source sweep, $G$ the number of GPUs, $C$ the cohort size, and $K$ the local steps per client. Batching is by audio duration, so batch count is a FLOP proxy; a shared server batch size $b_S$ is assumed. The per-GPU client load is $C\,K/G = 320$ batches for both pairs.}
\label{tab:fednst-compute}
\begin{tabular}{lllcc}
\toprule
 & & & \multicolumn{2}{c}{\textbf{Rel.\ compute/step}} \\
\cmidrule(l){4-5}
\textbf{Method} & \textbf{Server update} & \textbf{Batches/GPU} & \textbf{LS100$\to$LS860} & \textbf{LS$\to$CV} \\
\midrule
Static PL            & none                          & $0$              & -- & -- \\
Global/Online (ours) & $k$ supervised steps          & $k$              & $1.00$ & $1.00$ \\
Rao~\etal            & rehearsal on $\rho C$ clients & $\rho\, n_S / G$ & $1.00$ & $1.11$ \\
FedNST               & full source sweep             & $n_S / G$        & $1.25$ & $2.34$ \\
\bottomrule
\end{tabular}
\end{table}

\subsection{Server-Update Compute}
\label{app:compute}

Both client training and the server update are sharded across the $G$ GPUs, so the relevant quantity is the \emph{per-GPU} batch count: each round a GPU processes $C\,K/G$ client batches ($K$ local steps per client) plus its share of the server update, at a cost of $(\#\,\text{batches})\times 3 b_S$ for a common batch size $b_S$.
Methods differ only in the server term (\Cref{tab:fednst-compute}).
FedNST sweeps the \emph{entire} labeled source, $n_S/G = |\serverD|/(b_S G)$ batches per GPU; Rao~\etal rehearse $\rho C$ clients, $\rho\,n_S/G$; and our global/online teachers take $k$ data-parallel supervised steps ($k$ per GPU, independent of $G$).
Because client and FedNST-server work both scale as $1/G$, the relative slowdown is GPU-count-independent for a fixed pair.

Crucially, FedNST's server term grows with the \emph{source} size, while every other method's stays tied to the cohort: the per-GPU client load is $320$ batches for both pairs, but FedNST adds $n_S/G = 1500/16 \approx 94$ for LS100 $\rightarrow$ LS860 ($100$\,h source) versus $28800/64 = 450$ for LS $\rightarrow$ CV ($960$\,h source), while our teachers add only $k$.
A FedNST step therefore costs $\approx 1.25\times$ (LS100 $\rightarrow$ LS860) and $\approx 2.34\times$ (LS $\rightarrow$ CV) a global/online step (\Cref{tab:fednst-compute}).
All runs share a $7$-day wall-clock limit.
The other methods reach $2000$ FL steps for LS100 $\rightarrow$ LS860 and $3000$ for LS $\rightarrow$ CV within it, whereas FedNST's higher per-step cost prevents it from completing these budgets within the limit, so its \Cref{tab:min-wer-global-vs-online} entries reflect truncated runs.

\section{Quantifying Domain Shift}
\label{app:domain-shift}

The pseudo-label-source guideline in \Cref{subsec:guidelines} turns on whether the server and client data are in-domain (highly overlapping) or cross-domain. That judgment is usually made from domain knowledge; here we make it quantitative with a speaker-level similarity measure between two corpora, adapting the embedding-based data-selection representation of \citet{aldeneh2026datamatter}.

\paragraph{Per-speaker embeddings.}
Following \citet{aldeneh2026datamatter}, we represent each speaker by a single $1344$-dimensional vector formed by concatenating three per-utterance embeddings and averaging over that speaker's utterances: a $192$-dim speaker-verification embedding~\citep{desplanques2020ecapa} (voice timbre and speaker identity), a $768$-dim WavLM~\citep{chen2022wavlm} embedding (phoneme-level acoustic content), and a $384$-dim sentence-BERT~\citep{reimers2019sbert} embedding of the transcript (lexical and topic content).

\paragraph{Pairwise similarity.}
Given per-speaker embedding matrices $E_A \in \mathbb{R}^{N_A \times 1344}$ and $E_B \in \mathbb{R}^{N_B \times 1344}$ for corpora $A$ and $B$, we L2-normalize each row and, for each speaker $a \in A$, take its nearest neighbor in $B$, $s_a = \max_{b \in B} \cos(e_a, e_b)$.
The $A \rightarrow B$ score is the mean of $s_a$ over speakers in $A$, and we symmetrize a pair by averaging $A \rightarrow B$ and $B \rightarrow A$.
Higher values indicate more similar speaker populations, i.e., smaller domain shift.
The three modalities are concatenated without per-modality normalization, so the score is dominated by the speaker and semantic components.

\paragraph{Results and use as a guideline.}
\Cref{tab:domain-shift} reports the measure for the corpora we study.
The two in-domain pairs (LS100 $\leftrightarrow$ LS860 and CV10 $\leftrightarrow$ CV90) score $0.56$ and $0.59$, both well above the cross-domain pairs ($0.38$--$0.44$).
The gap from the lowest in-domain score ($0.56$) to the highest cross-domain score (LS $\leftrightarrow$ Ted, $0.44$) exceeds the spread among the cross-domain pairs themselves.
The measure therefore separates in-domain from cross-domain and can serve as a quantitative proxy for the server--client overlap that the guideline in \Cref{subsec:guidelines} uses to choose the teacher: a high score indicates an in-domain pair (favoring the online teacher), and a low score a cross-domain pair (favoring the more robust global teacher).
Being symmetric, the measure captures the overall distance between two speaker populations rather than the direction of transfer; the direction-asymmetric instability of \Cref{sec:server-update} is governed instead by the dispersion of the server data.

\begin{table}[H]
\centering
\small
\caption{Speaker-level similarity between corpora, computed as the symmetrized nearest-neighbor cosine over the $1344$-dim concatenated embeddings. Higher means more similar speaker populations, i.e., smaller domain shift. The in-domain pairs (top) score well above the cross-domain pairs (bottom).}
\label{tab:domain-shift}
\begin{tabular}{lc}
\toprule
\textbf{Pair} & \textbf{Similarity} \\
\midrule
LS100 $\leftrightarrow$ LS860 (in-domain) & $0.56$ \\
CV10 $\leftrightarrow$ CV90 (in-domain)   & $0.59$ \\
\midrule
LS $\leftrightarrow$ Ted  & $0.44$ \\
LS $\leftrightarrow$ CV   & $0.40$ \\
Ted $\leftrightarrow$ CV  & $0.38$ \\
\bottomrule
\end{tabular}
\end{table}

\section{Contributions}
The overall vision for studying semi-supervised learning for federated learning in ASR was conceived by Wonho Bae, Sheikh Shams Azam, Tatiana Likhomanenko, Martin Pelikan, and Jan “Honza” Silovsky, who identified the gap in current research and defined the problem scope. This work builds on motivating techniques from prior co-authored works published by the group: (i) private federated learning for ASR \cite{pelikan2023dp_fl_asr, azam2023fl4asr} led by Martin Pelikan, Sheikh Shams Azam and Tatiana Likhomanenko  and (ii) Iterative Pseudo-Labeling \cite{xu2020ipl, aldeneh2025speakeripl} led by Zakaria Aldeneh. Specific contributions of the authors can be attributed as:
\begin{itemize}[leftmargin=4mm]
    \item \textbf{Algorithm Design.} The design of the algorithm and ablations was led by Wonho Bae, Jan “Honza” Silovsky, Tatiana Likhomanenko, and Sheikh Shams Azam  in consultation with Martin Pelikan and Zakaria Aldeneh.
    \item \textbf{Implementation and Experimental Results.} Wonho Bae developed the Semi-Supervised FL for ASR training pipeline by adapting the PFL codebase from Martin Pelikan and Tatiana Likhomanenko, conducted the domain shift analysis in \Cref{app:domain-shift} using code and guidance from Zakaria Aldeneh, and led all experiments. The code and data analysis were further reviewed by Tatiana Likhomanenko and Sheikh Shams Azam. Sheikh Shams Azam also contributed to the evaluations and ablation studies. All work was done in consultation with the other authors.
    \item \textbf{Writing and Paper Preparation.} The manuscript was written by Wonho Bae and Sheikh Shams Azam. It was edited and reviewed by all other authors.
\end{itemize}
%% <<<<< end body.tex

\applefootnote{\textcolor{textgray}{\sffamily Apple and the Apple logo are trademarks of Apple Inc., registered in the U.S. and other countries and regions.}}

\end{document}